\documentclass[10pt,journal,compsoc]{IEEEtran}
\ifCLASSOPTIONcompsoc
  \usepackage[nocompress]{cite}
\else
  \usepackage{cite}
\fi
\ifCLASSINFOpdf
\else
\fi
\usepackage{graphicx}
\usepackage{subfigure}
\usepackage{chngcntr}
\usepackage{booktabs}
\usepackage{multirow}
\usepackage{diagbox}

\begin{document}
%
\title{Energy-based Periodicity Mining with Deep Features for Action Repetition Counting in Unconstrained Videos}
%
%
%

\author{ Jianqin~Yin Yanchun~Wu
       Huaping~Liu Yonghao Dang Zhiyi Liu
        and Jun Liu
\IEEEcompsocitemizethanks{\IEEEcompsocthanksitem Jianqin~Yin, Yanchun Wu, Yonghao Dang, and Zhiyi Liu are with Automation School of Beijing University of Posts and Telecommunations, 100876, Beijing, China.\protect\\
 Huaping Liu is with Department of Computer Science, Tsinghua University, Beijing, China. \protect\\
Jun Liu is with Department of Mechanical Engineering, City University of Hong Kong, Hong Kong, China.\protect\\
Jianqin Yin and Yanchun Wu are contributed equally to this work.  Jianqin Yin and Jun Liu are the corresponding authors.

}
\thanks{Manuscript received Nov 19, 2019}}

\IEEEtitleabstractindextext{%
\begin{abstract}
Action repetition counting is to estimate the occurrence times of the repetitive motion in one action, which is a relatively new, important but challenging measurement problem. To solve this problem, we propose a new method superior to the traditional ways in two aspects, without preprocessing and applicable for arbitrary periodicity actions. Without preprocessing, the proposed model makes our method convenient for real applications; processing the arbitrary periodicity action makes our model more suitable for the actual circumstance. In terms of methodology, firstly, we analyze the movement patterns of the repetitive actions based on the spatial and temporal features of actions extracted by deep ConvNets; Secondly, the Principal Component Analysis algorithm is used to generate the intuitive periodic information from the chaotic high-dimensional deep features; Thirdly, the periodicity is mined based on the high-energy rule using Fourier transform; Finally, the inverse Fourier transform with a multi-stage threshold filter is proposed to improve the quality of the mined periodicity, and peak detection is introduced to finish the repetition counting. Our work features two-fold: 1) An important insight that deep features extracted for action recognition can well model the self-similarity periodicity of the repetitive action is presented. 2) A high-energy based periodicity mining rule using deep features is presented, which can process arbitrary actions without preprocessing.  Experimental results show that our method achieves comparable results on the public datasets YT\_Segments and QUVA.
\end{abstract}

}

\maketitle

\IEEEdisplaynontitleabstractindextext

%
\IEEEpeerreviewmaketitle

\IEEEraisesectionheading{\section{Introduction}\label{sec:introduction}}

\begin{center}
\begin{figure}
\centering
\includegraphics[width=3.3in]{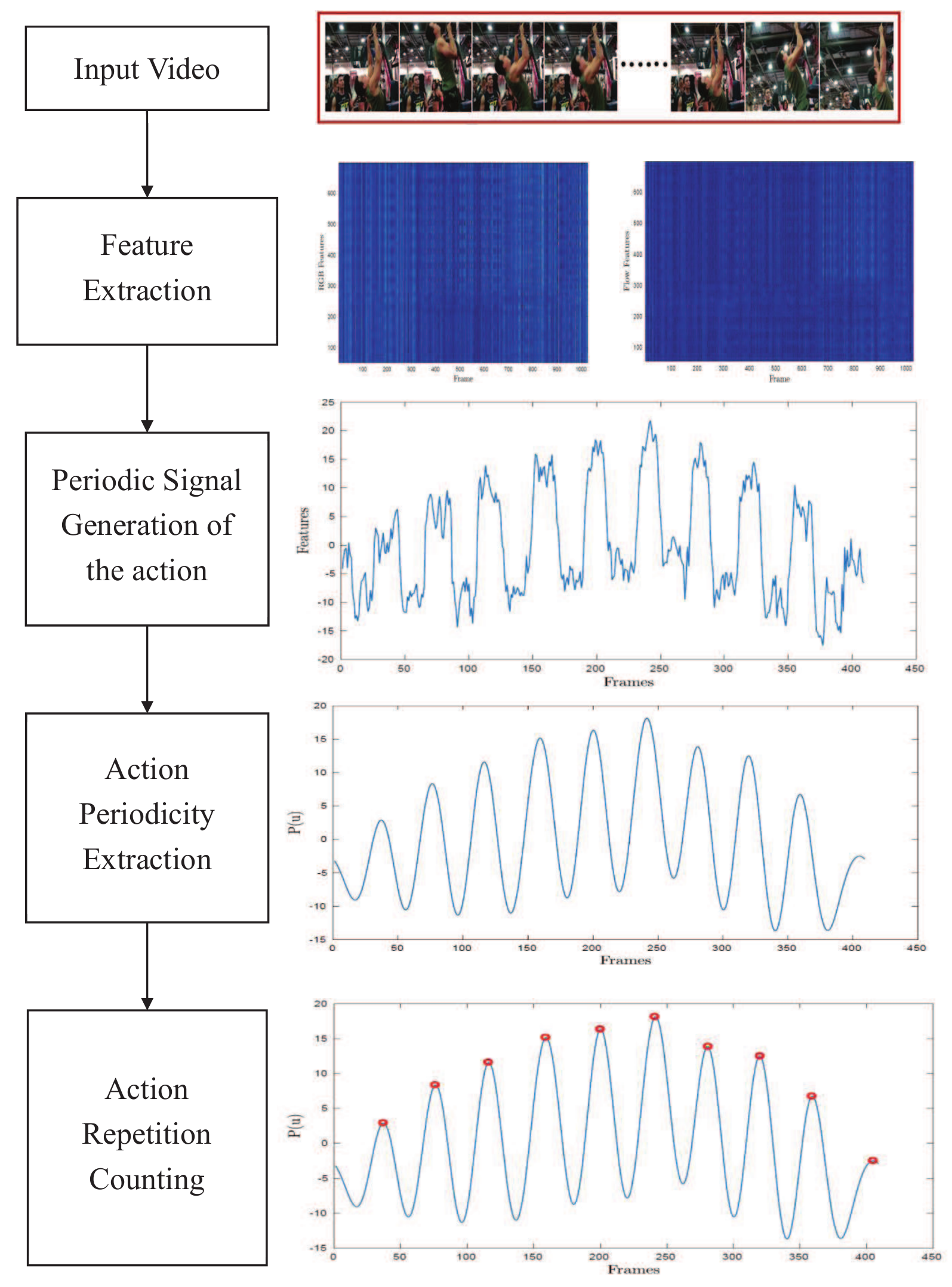}
\caption{ The framework of action repetition counting in this paper. }
\label{figsv}
\end{figure}
\end{center}

%
%
%
%
\IEEEPARstart{V}isual action repetition in real life appears in many applications, such as sports, music playing and manufacturing assembly. It is important to count the repetition of a specific motion in videos, which can be used for video question answering\cite{ref0}, action classification \cite{ref23} \cite{ref8} \cite{ref15}, segmentation \cite{ref2} \cite{ref3} \cite{ref4}, 3D reconstruction \cite{ref9} \cite{ref12}, motion tracking \cite{ref1} and motion planning of robots \cite{ref18}. Due to the diversity of motion patterns and the limitations in video capturing(e.g.,camera movement), the development of an universal solution for counting the repetitive actions remains under-explored.\\
 To count the action repetition,  early methods usually assumed that repetitive motions occurred in fixed scenes with regular periodicity. With this assumption, traditional features were used to analyze the action repetition, including human skeleton obtained by sensor device \cite{ref9} \cite{ref12}, the wavelength spectrum \cite{ref6}. However, the actions to be counted are usually captured in complex dynamic scenes and have variable periodicity over different periods, making the traditional features not suitable for the counting tasks. To tackle this complexity involved in real circumstances, two methods have been proposed in recent years. In \cite{ref21}, multiple repetitive motion modes are simulated to construct the periodicity of the repetitive actions to realize the counting. Because the simulated motion modes are fixed, the algorithm can handle the specified modes well. But the performance significantly decreases for actions with other repetitive modes \cite{ref22}. In \cite{ref22}, a counting method based on the detection of the moving area is proposed to achieve an improved performance. However, this method relies on additional preprocessing steps to detect the moving area. In sum, the methods based on the simulated action modes are not effective to count the varied types of repetitive motions, and the moving-area methods are highly dependent on the preprocessing performance. This motivates us to find an action repetition counting scheme that can work for unspecified motion modes without relying on extra detection steps.  \\
There are many challenges in action repetition counting for unconstrained videos. In the unconstrained videos, besides the interested repetitive action, there exists other motion information such as changes in the background, actions of the false objects and other unrelated movements. How to distinguish the periodic actions from these various irrelevant signals is prerequisite problem to address for counting the repetitive action. Moreover, action repetition modes are very different in different actions. For example, the repetition mode can be rotation, swinging, translation, and other modes. How to discover the relationship between the periodicity and various repetition modes is the another challenge. Additionally, the amplitude and the frequency of the repetition within the same action can also be different. Therefore, even for the same action, how to discover the features to count the repetition is an additional challenge. \\
Although the repetition counting is very important,  because of the aforementioned difficulties, the research progress on it is relatively slow in recent years. In contrast, great advances have been achieved by using deep learning methods for action recognition  \cite{ref26} \cite{ref32} \cite{ref33}, which opens up new possibilities to propose new solutions for repetition counting. It has been proven that deep ConvNet can capture versatile and robust action features for action recognition, which illustrates that the deep features contain rich information of the action.  For repetition counting, there is an important clue that the self-similarity periodic dynamics is the key. Accordingly, we propose in this paper that the deep features are helpful for mining the self-similarity periodicity of the action and can be used to count the repetition. This insight relieves us from the preprocessing and helps in addressing the challenges of modeling the periodicity of the unconstrained videos.
\\
Although deep features may include periodic rules, the high dimension of the feature space makes it difficult to discover the self-similarity modes.  In order to mine the repetitive rules from the deep features, we use Principal Component Analysis (PCA) to reduce the high dimension of the deep features. And we found that the non-stationary repetitive signals frequently appear in the first-dimensional principal component. This is an important insight for modeling the action repetition. Using PCA, the high dimensional deep features can be converted to a one-dimension waveform. In a word, the periodic signal can be generated combining deep features and PCA.  \\
For the video including the repetitive action, most of the energy of the video usually comes from the repetitive motions due to the repetition. From this perspective, once we obtain the waveform that contains the periodic motions, we can locate the periodic motion using the highest energy rules. In detail, frequency analysis is used to locate the signals with highest energy corresponding to the repetitive action automatically, and then we can finish counting based on the located signal.\\
 The framework of our method is shown in Fig. \ref{figsv}. Our framework includes four steps. At first, deep features are extracted using two-stream deep ConvNets, generating spatial and temporal features separately. Secondly, one-dimensional periodic signal is generated based on the deep features using PCA. Corresponding to the spatial and temporal features, there are two one-dimensional periodic signals. For simplicity,  we only show the periodic signal of the spatial features in Fig. \ref{figsv}. Thirdly, the periodicity of the repetitive action is mined based on the highest energy rules using Fourier transform, filtering and inverse Fourier transform. Finally, peak detection is used to finish the repetition counting.

In this paper, a new action repetition counting method is proposed for solving the unconstrained action repetition counting in videos, and main contributions are summarized as follows.
(1) We propose an energy-based action repetition counting method without extra preprocessing, which can be used to effectively count the repetition of the action with arbitrary periodic motions for unconstrained videos.
(2) We give an important insight that the periodic self-similarity movement information can be well modeled by the deep features of the action used in action recognition.
(3) We propose the Fourier transform with Multi-stage threshold Filter (FMF) scheme to automatically mine the interesting repetitive action and remove noises to improve the repetition counting performance.\\
The remainder of the paper is organized as follows. The next section investigates the related work. Section III discusses our algorithm in detail. The datasets and evaluation criteria are described in Section IV. Experimental results and analysis are illustrated in Section V. We conclude our paper in Section VI.




\section{Related work}
Action repetition counting is usually realized by converting the video into a one-dimensional waveform with the repetitive motion structures \cite{ref9} \cite{ref12} and then analyzing the spectral or frequency component by Fourier transform or wavelet analysis. Waveform analysis is also widely used in the periodic movement analysis \cite{ref5} \cite{ref10}, 
which is very related to the repetition counting. At the same time, as discussed in the Introduction, the action feature is another important problem for counting. Therefore, we will review the related work from the following two aspects: periodic movement analysis and deep features for action analysis.\\
\subsection{Periodic movement analysis}
 The existing methods have achieved remarkable results in video action periodic analysis tasks. Burghouts et al. \cite{ref6} proposed a spatiotemporal filter bank for estimating of action repetition, which could work online. But it was limited to the motion of stationary scenes, and the filter bank was manually adjusted. Laptev et al. \cite{ref7} used a matching method for action counting, whose primary work is to detect and segment repetitive motions using the geometrical constraints generated by the same motion repeatedly when the viewpoint changes. Ormoneit et al. \cite{ref1} used functional analysis to represent cyclic movement. Ribnick et al. \cite{ref9} \cite{ref24} found that it is possible to reconstruct accurately periodic movements in 3D from a single camera view. Based on this research, they applied 3D reconstruction to gait recognition. Ren et al. \cite{ref19} and Li et al. \cite{ref20} developed two autocorrelation counting systems based on matching visual descriptors. Although both systems completed the repetition counting, they are postprocessing methods, which are only applicable to specific domains of restricted video. Pogalin et al. \cite{ref5} get the information of a certain part of the body by tracking the object of interest, then performs pPCA and spectral analysis followed by detection and frequency measurement. But its purpose is to estimate the degree of periodic motion but not to count the repetition. Based on the human skeleton points captured by Kinect, Wang et al. \cite{ref4} proposed an unsupervised repetitive motion segmentation algorithm based on the frequency analysis of the motion parameter, zero-velocity cross detection and adaptive $k$-means clustering. Although the above methods lay the foundation for the video repetitive counting task, they only realized the simple repetitive motion estimation of a fixed scene and had a poor performance on the diversity and non-stationary motion, which commonly exist in the real applications.\\
In recent years, Kumdee et al. \cite{ref8} used the image self-similarity measure as the input of the multi-layer the perceptron neural network to determine whether the input video is a repetitive action. This method is relatively stable to image changes, noise, and low-resolution images. However, they focus on classifying that the video is a repetitive video or not but not counting. Levy et al. \cite{ref21} proposed a method to count the repetitive action of the videos using the convolutional neural network. They used synthetic data to simulate four motion types for the periodic motion and carried out network training and prediction. In the test, the region of interest was calculated through the motion threshold for the test data. The motion cycle was classified through the classification network to complete the repetitive counting task. The method showed excellent performance on YT\_Segments dataset. However, their algorithm decayed a lot when there are actions with different repetitive modes from the trained modes. The wavelet transform was presented in \cite{ref22} to better deal with more complex and diverse video dynamics. From the flow field and its differentials, they derived 18 totally different repetitive perception. Based on the gradient, curl and divergence, a motion foreground segmentation representation based on flow was realized, and remarkable results were obtained. However their methods needs the foreground segmentation, which is also a difficult problem. Therefore, we propose a method without extra preprocessing. \\
\subsection{Deep features for action analysis }
 CNNs have been widely used in action recognition. 
 Some of these CNNs use deep architectures with 2D convolutions to extract translation-invariant features in the video frames \cite{ref26}. Specifically, Karpathy et al. \cite{ref26} first introduced a CNN based method for action recognition and organized a large-scale sports video dataset (i.e., Sports-1M dataset) for training deep CNNs. To model the temporal information of the action, two-stream based CNN learning framework \cite{ref32} \cite{ref33} has been proposed. The two streams mean the spatial stream represented by RGB values and the temporal stream represented by pre-computed optical flow features. Because of the excellent balance between efficiency and effectiveness, the BN-Inception Network \cite{ref31} is used as the backbone of the framework. The prominent characteristic of BN-Inception network is the Inception module, which carries out multi-scale processing and fusion of image features to extract better feature representation. Moreover, it is well known that the 3*3 convolution kernel has the best performance in VGG \cite{ref28}, and a very effective Batch Normalization(BN) method has been proposed to accelerate the learning of data distribution during training, making the accuracy of the classification improve significantly. In addition, the deep ConvNets can take pictures of any form as input to extract features. The training of deep ConvNets requires a large number of training samples to achieve good performance in action modeling. Nowadays, a large number of publicly available video datasets provide great convenience. Therefore, we extract the deep features of our experimental data using BN-Inception Network in this paper.

\begin{center}
\begin{figure}
\centering
\includegraphics[width=3.5in]{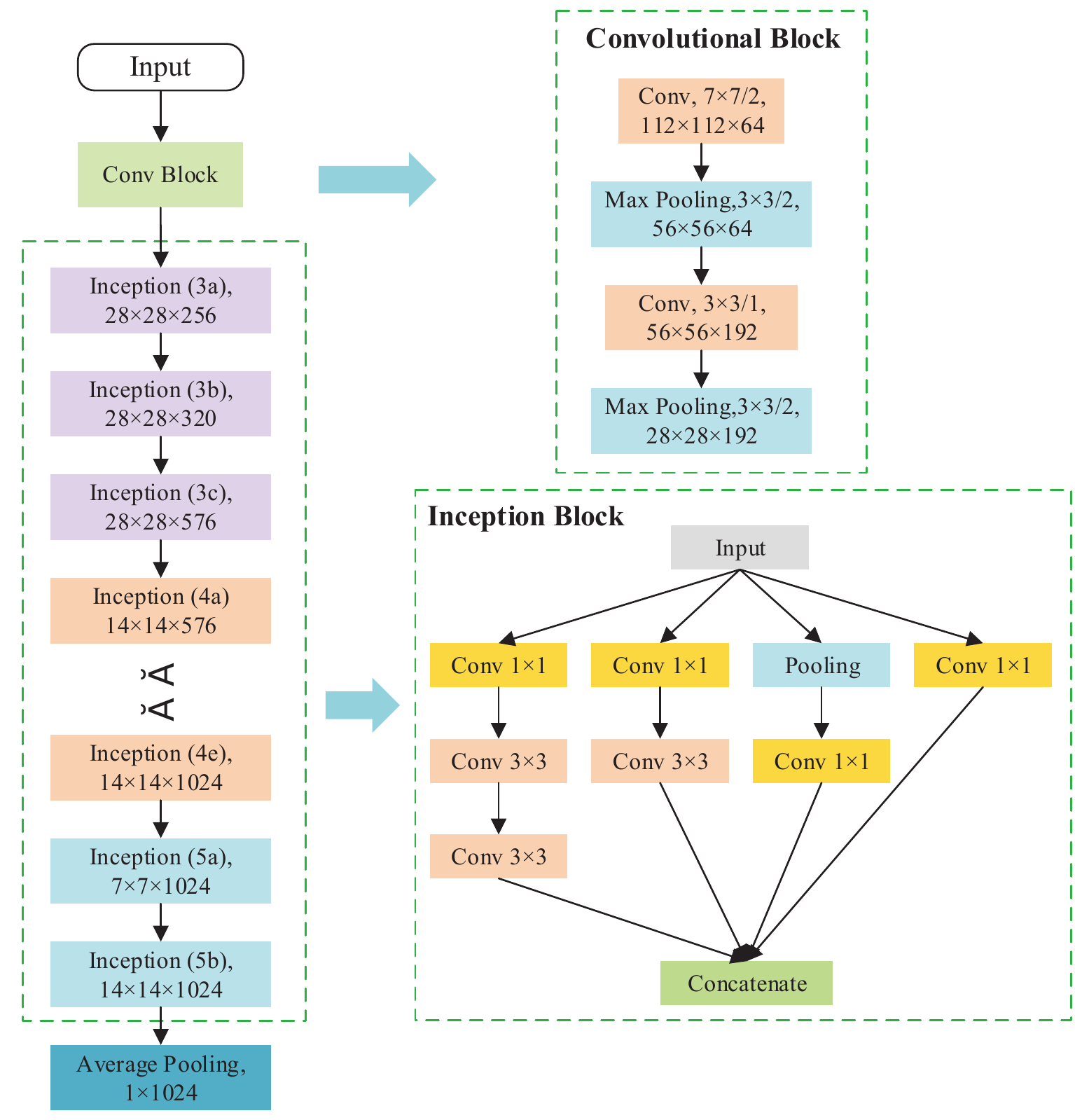}
\caption{ The framework of BN-Inception Network}
\label{fignet}
\end{figure}
\end{center}

\section{Action Repetition Counting}
In this part, we will discuss the proposed algorithm. As shown in Fig. \ref{figsv}, our algorithm includes four steps. Firstly, deep features of the unconstrained videos are extracted using deep ConvNets. Secondly, based on the high-dimensional deep features, the periodic signal is generated using PCA, and can obtain a one-dimensional waveform reflecting the repetitive changes of the videos. Thirdly, the action repetition rules are extracted based on the highest energy rules extracted from the 1D waveform, combining Fourier analysis, a new proposed multi-stage threshold filter, and the inverse Fourier transform.  Finally, using the waveform, peak detection is used to count the repetition.
\subsection{Deep Feature Extraction based on BN-Inception ConvNets}
The Inception v2 based on Batch Normalization network \cite{ref26} is used to obtain the features of the action, as shown in Fig. 2. It was trained on the large public Kinetics dataset \cite{ref11}, which contains 300,000 clip videos from real scenes, including 400 action categories, which is the largest action recognition dataset by far. The model we used is the best network trained in \cite{ref28}.\\
Two networks are used to extract robust action features, operating on two components, spatial and optical flow separately. The spatial flow network operates on the RGB image, which extracts spatial features describing the scene and object information. The optical flow network takes the pre-computed optical flow images as the input to extract the temporal features, which describe the motion information of the video. Robust spatiotemporal features are extracted by this method. Fig. 3 is a structure for feature extraction of the action.

\begin{center}
\begin{figure}
\centering
\includegraphics[width=3.3in]{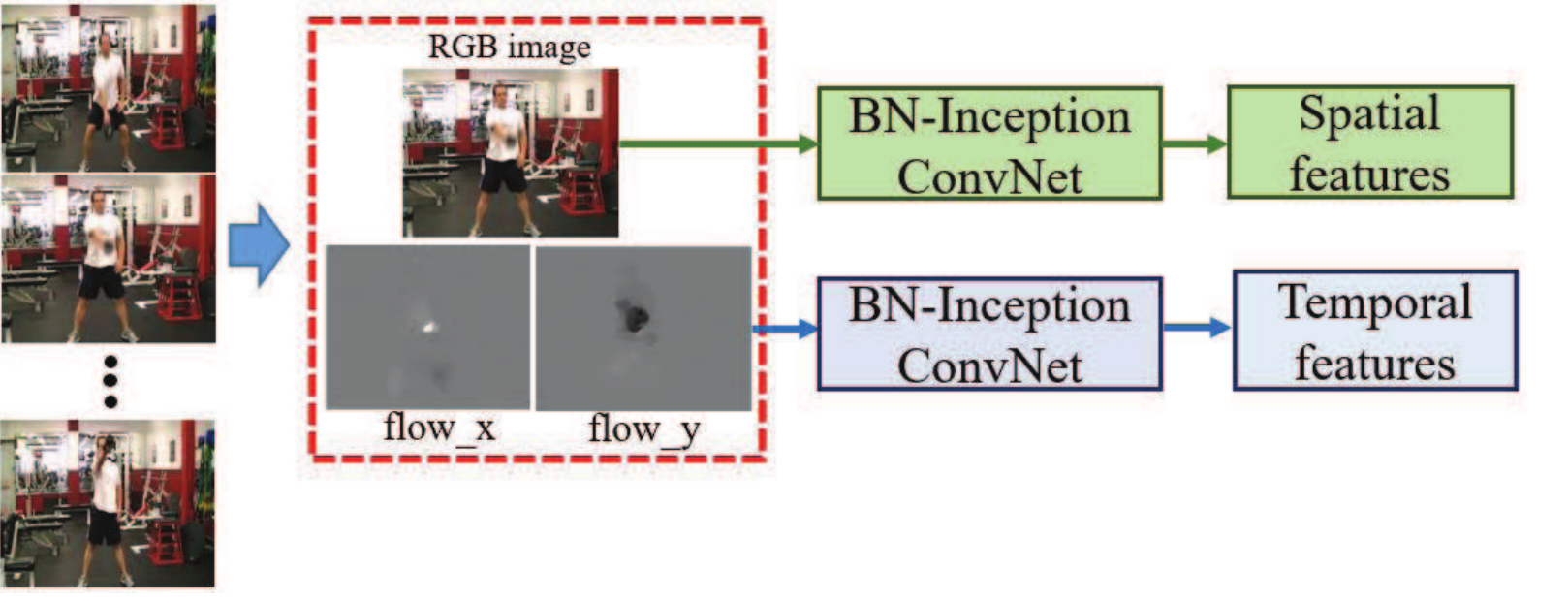}
\caption{ The structure for spatiotemporal features extraction. The input modalities of BN-Inception are the RGB images and optical flow fields (x, y directions).}
\label{fignet}
\end{figure}
\end{center}

Clipping and rotation of training images, to decrease the influence of the noise and increase the stability of features, are used to get the image set. In the process of feature extraction, we take the image set as network input and get the features in the $avg\_pool$ layer. Then the summation and the average features are computed in each dimension. Finally, the spatial features, denoted by $Static_{fea}$, and temporal features, denoted by $Dynamic_{fea}$ are extracted for the single image, respectively. Due to the designed structure of the network, the dimension of the feature vector is 1024, as shown in equations (1) and (2).

\begin{small}\begin{equation}Static_{fea}=(s_1,s_2,\ldots,s_{1024})\end{equation}\end{small}\label{eq:staticf}

\begin{small}\begin{equation}Dynamic_{fea}=(d_1,d_2,\ldots,d_{1024})\end{equation}\end{small}\label{eq:dynamicf}

\begin{center}
\begin{figure}
\centering
\setcounter{subfigure}{0}
\subfigure[RGB features of the action video]{\label{RGBf}\includegraphics[height=1in,width=0.25\textwidth]{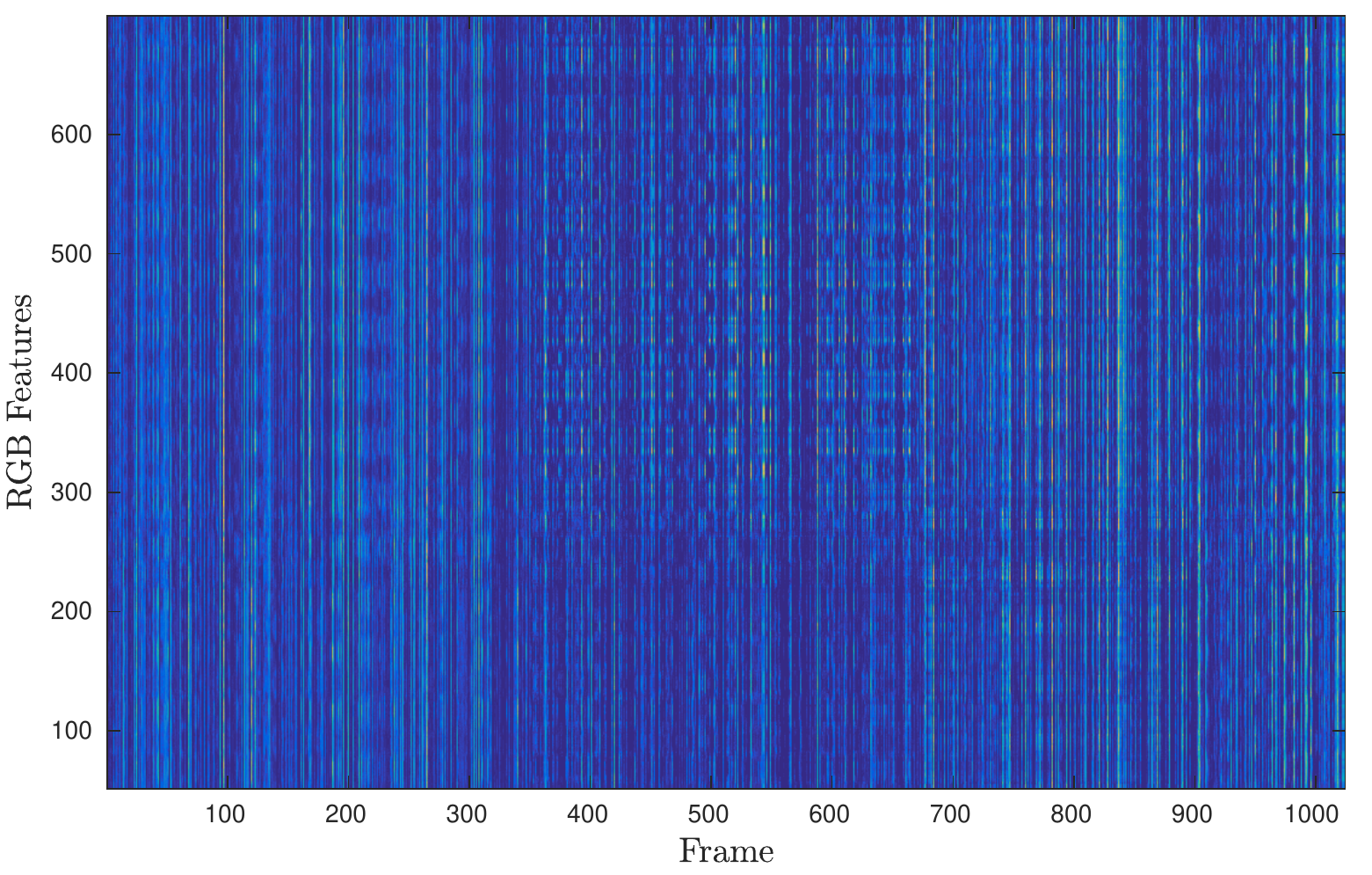}}
\subfigure[Optical features of the action video]{\label{optf}\includegraphics[height=1in,width=0.25\textwidth]{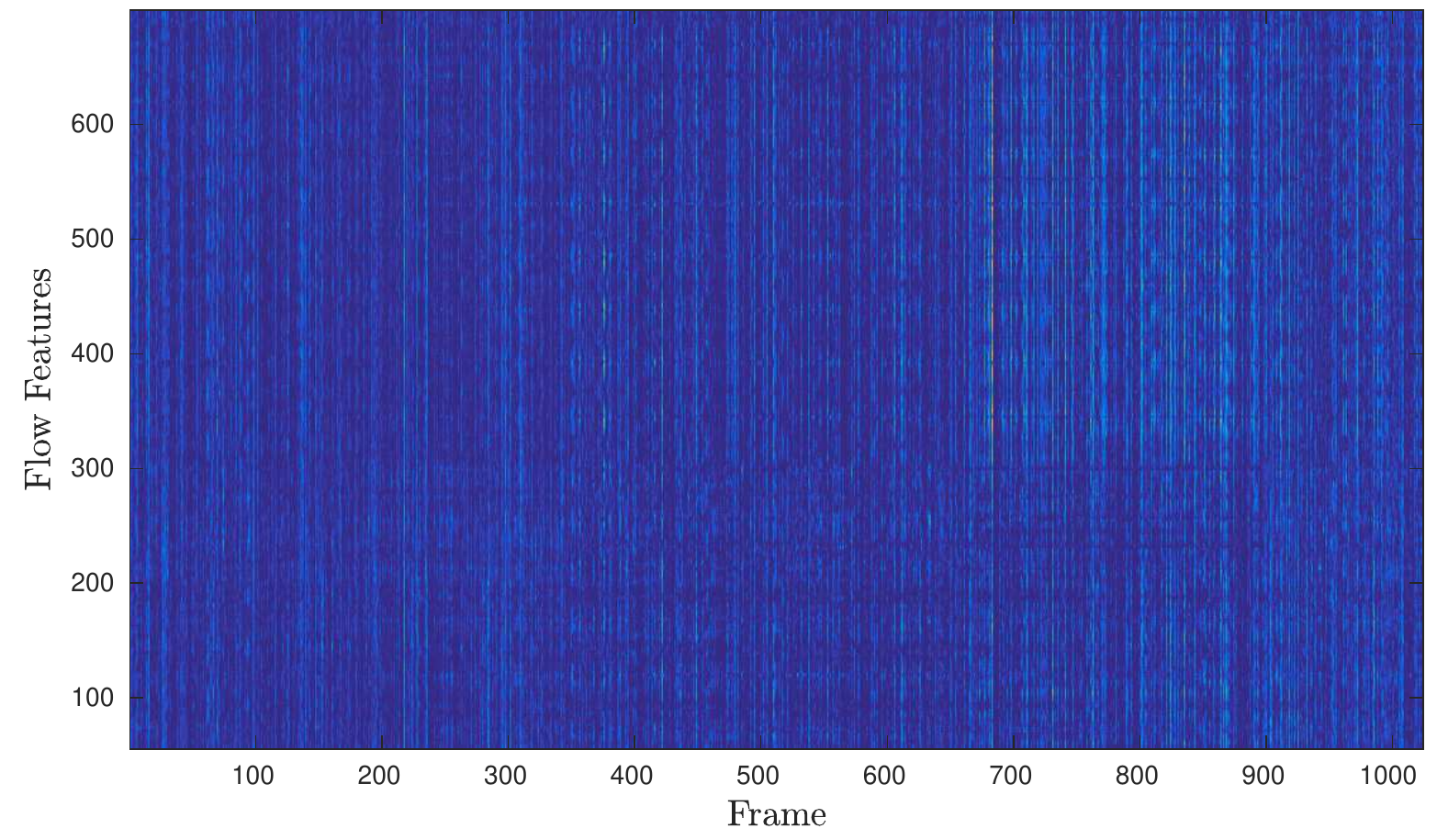}}

\caption{The visualization of deep features}
\label{fig:depF}    
\end{figure}
\end{center}

 As discussed above, deep features can be obtained using the pre-trained models and we do not need to retrain the model using repetitive actions. The extracted deep features of the video is visualized in Fig. \ref{fig:depF}, where RGB features are given in Fig. \ref{RGBf} and optical flow features are given in Fig. \ref{optf}. From the visualization results, although we can find some specified patterns; it is difficult for us to find the periodic rules using this high-dimension features. Therefore, we need some other methods to mine the periodicity of the repetitive action.

\subsection{Periodic Signal Generation}
To extract the periodicity information, we mine the hidden periodic action rules from two different features, spatial features($Static_{fea}$) and temporal features($Dynamic_{fea}$). The mining method for these two features are the same; therefore, we take spatial features $Static_{fea}$ as an example to explain our extraction method. Although deep features can classify the actions well, the counting of the repetition of the action is totally different from the classification of the action. For classification, one action is considered as a whole, and it focuses on the difference between different actions. On the contrary, action repetition counting focuses on locating the repetition of the same motion pattern. Therefore, the deep features extracted directly for action recognition may not be suitable for counting. We transform the high-dimensional features into an intuitive waveform by extracting the primary component of its covariance matrix.\\
For simplicity, we represent the static features $Static_{fea}$ of the $i$th frame as $f_i=Static_{fea}$, where $i$ is the frame index ranging from 1 to $N$. Therefore, for every action video, its features can be represented by the features of all its frames, denoted by $F=[f_1,f_2,\ldots,f_N] $, where $F$ is a 2D matrix with dimensions of $N \times D$,  $N$ is the total number of the frames in a video, $D$ is the dimension of the spatial features, 1024. The detailed process is as follows.\\

 Firstly, for the feature matrix $F$, we preprocess its $i$th components using ${\bar f_i} = \frac{{{f_i}}}{D}$, where, $i=1,2\ldots,N$. Then we obtain the mean matrix according to the formula $\bar F = \left[ {{{\bar f}_1},{{\bar f}_2} \cdots {{\bar f}_N}} \right]$ . Then the transformation matrix $\hat F$ is computed using $\hat F = F - \bar F$. Finally, the covariance matrix is calculated according to equation (3).

\begin{center}
\begin{figure}
\centering
\setcounter{subfigure}{0}
\subfigure[The first 1D of the component of the optical flow features]{\label{fig4a}\includegraphics[height=1in,width=0.2\textwidth]{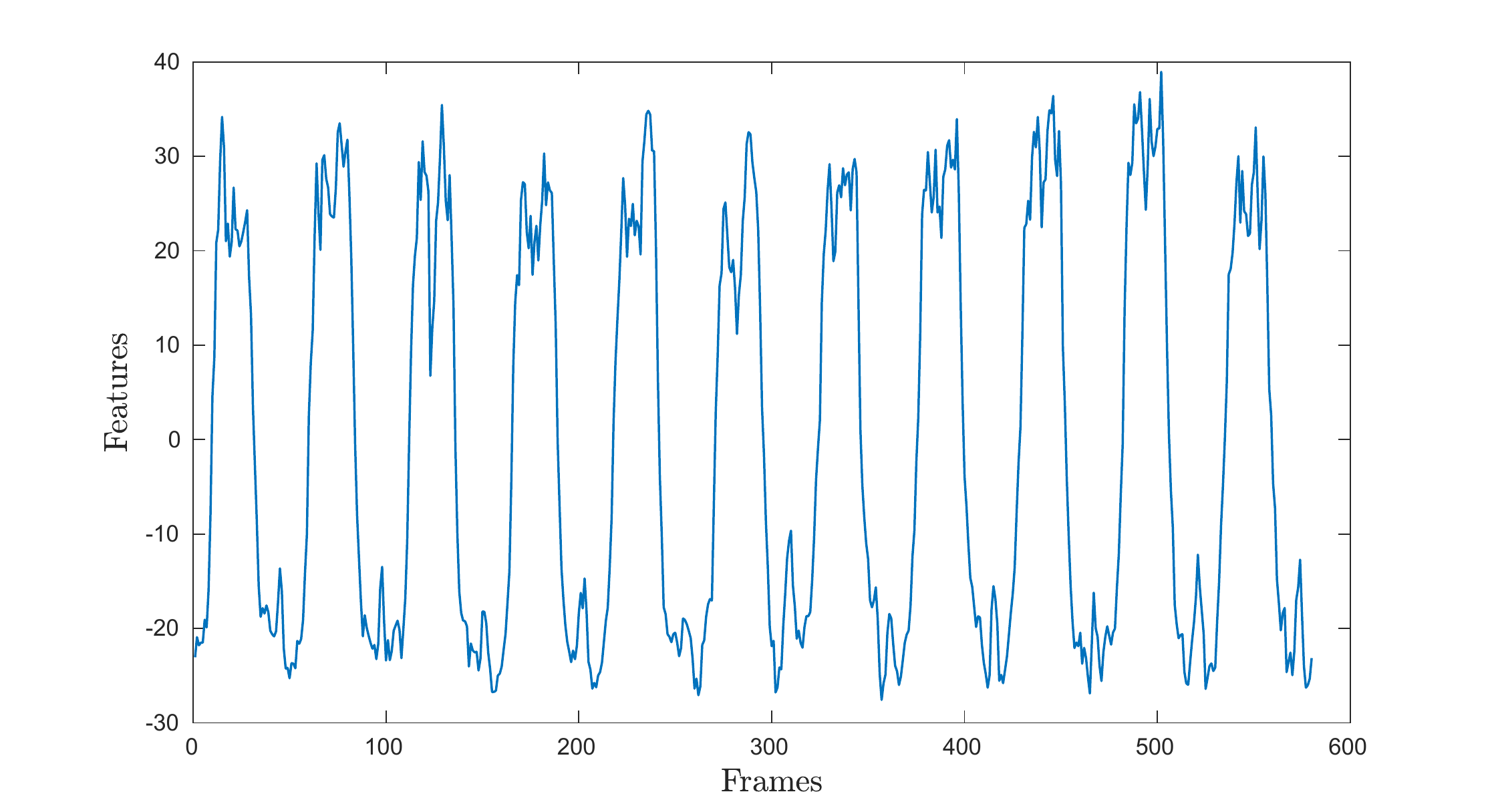}}
\subfigure[The first 10D of the component of the optical flow features]{\label{fig4b}\includegraphics[height=1in,width=0.2\textwidth]{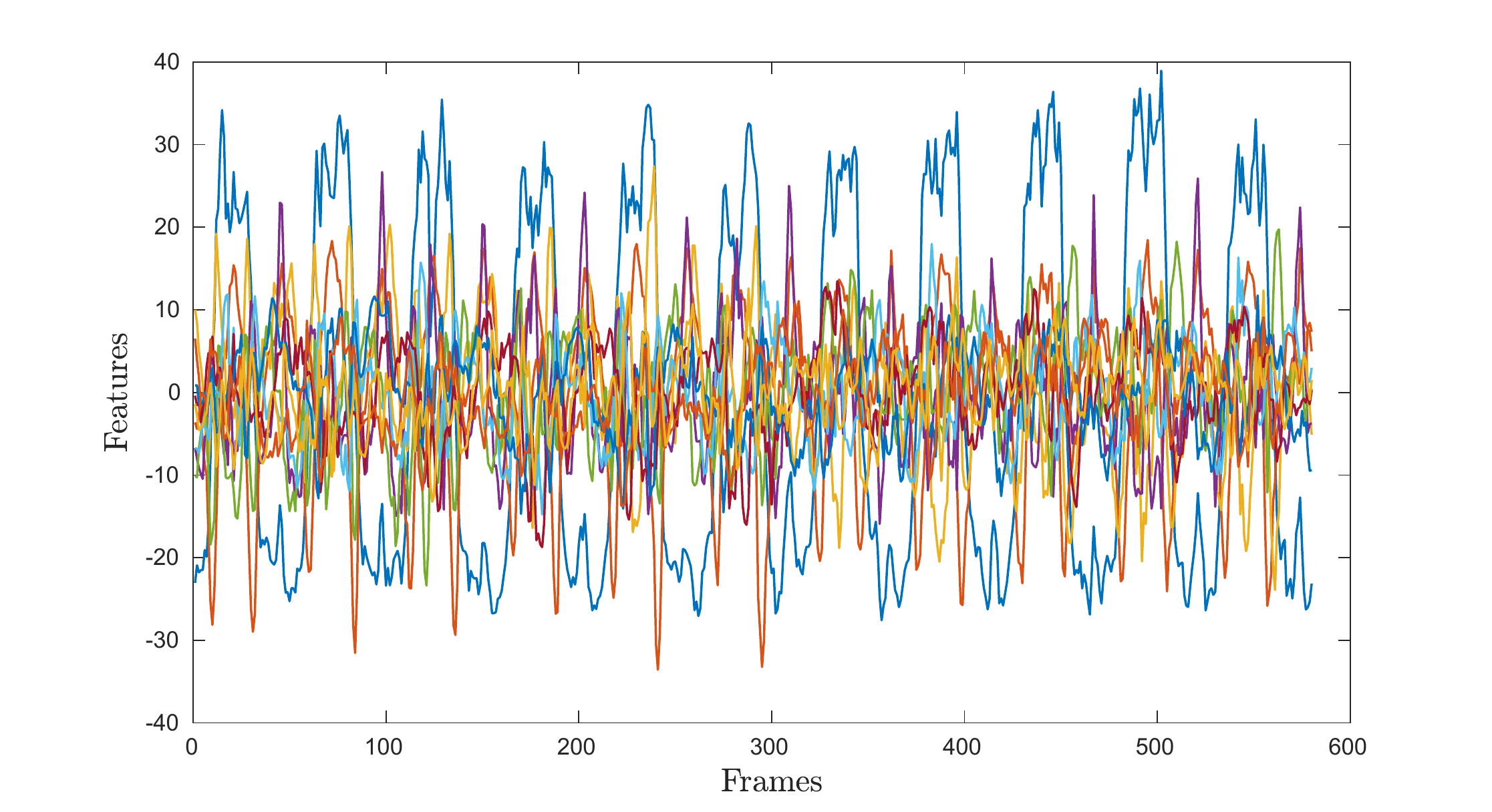}}
\subfigure[The first 1D of the component of the RGB features]{\label{fig4c}\includegraphics[height=1in,width=0.2\textwidth]{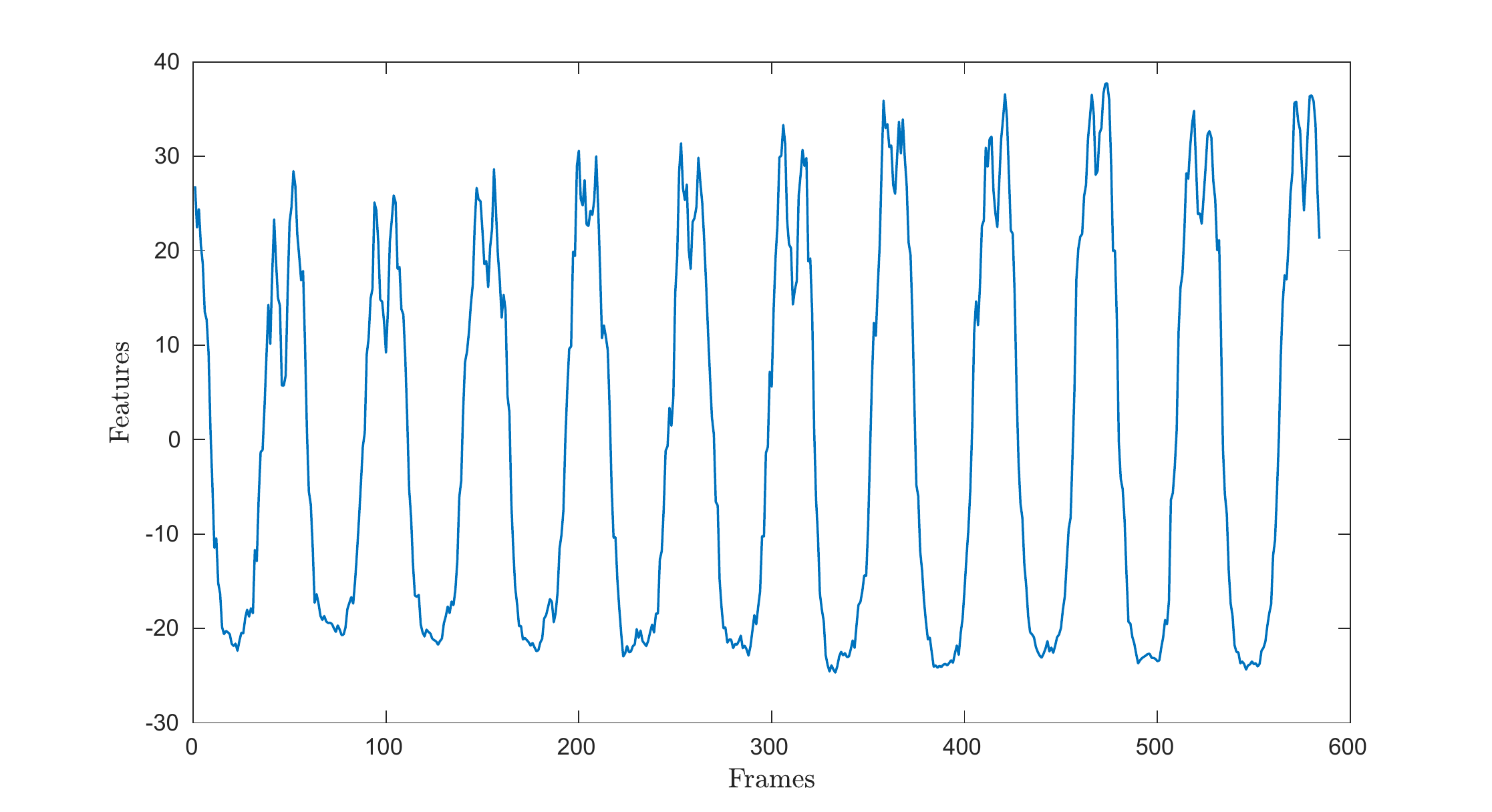}}
\subfigure[The first 10D of the component of the RGB features]{\label{fig4d}\includegraphics[height=1in,width=0.2\textwidth]{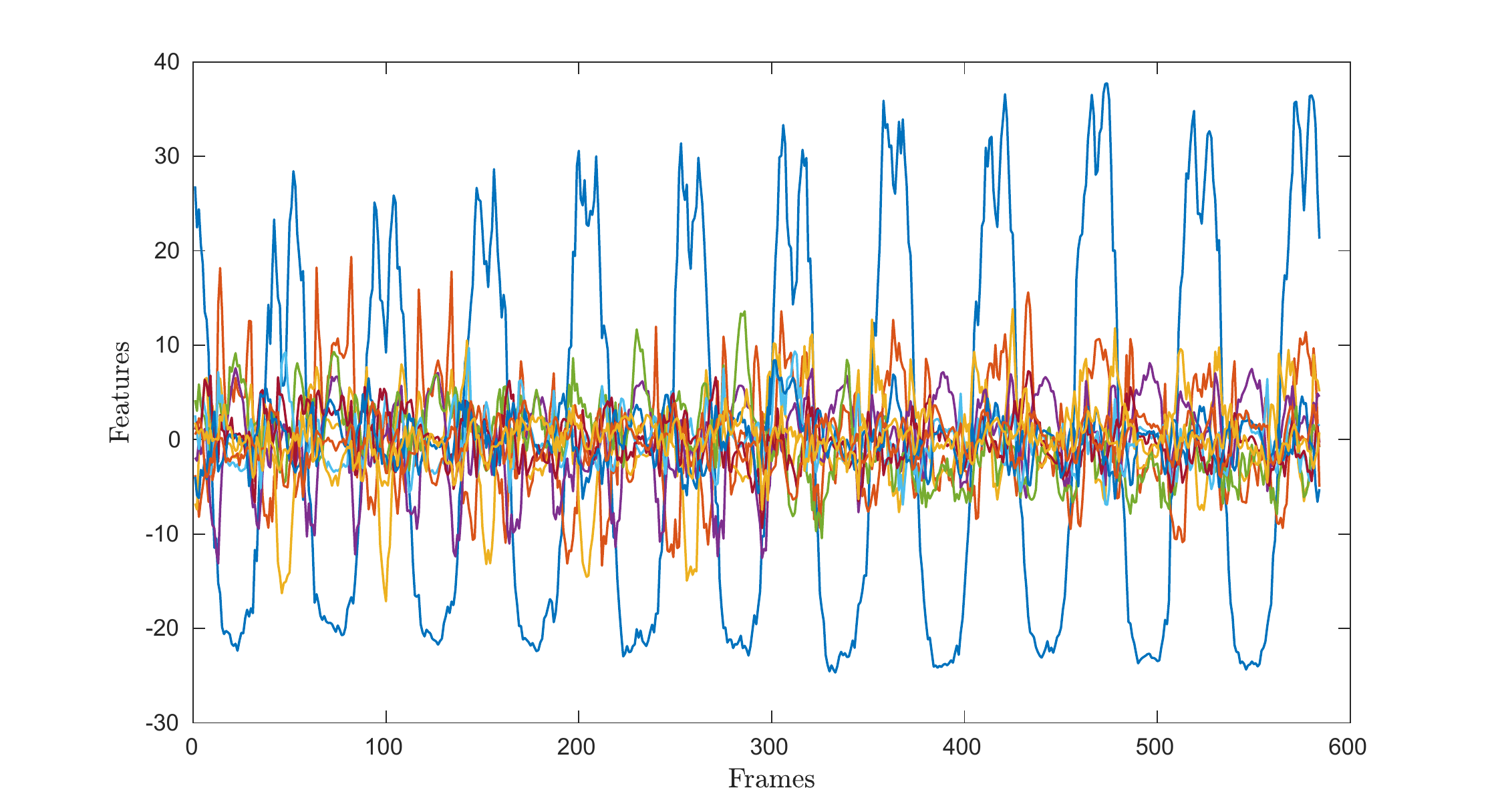}}

\caption{Periodicity visualization of the repetitive action}
\label{fig:PCAf}    
\end{figure}
\end{center}

\begin{small}\begin{equation}COV = \frac{1}{D}\hat F{\hat F^T} = V\Lambda {V^T}\end{equation} \end{small}
We also can compute the eigenvalues and eigenvectors of the covariance matrix. The corresponding results are separately denoted by their matrix form as $\Lambda  = diag\left( {{\lambda _1},{\lambda _2} \cdots  \cdots {\lambda _D}} \right)$ and $V = \left[ {{\mu _1},{\mu _2} \cdots {\mu _N}} \right]$, where each $\mu_i$ is a vector with $1 \times D$. We arrange $\Lambda$ according to the value of its eigenvalue from large to small. And according to the new order of eigenvalues, we rearrange $V$ to $ V'$ in columns. Then, we reserve the first eigenvector to get the transformation matrix $V_1'$. The size of $V_1'$ is $D \times 1$. Therefore the mapped matrix $P=(p_1,p_2,p_3\ldots p_N)$ can be computed according to formula (4), 
where $p_i$ is the principal component of $i$th frame of the videos, $i={1,2,\ldots N}$. The size of $P_i$ is $N \times 1$, by which the high-dimensional video features are transformed into the new space constructed by 1D waveform, as shown in Fig. \ref{fig:PCAf}.


\begin{equation}{p_{_i}} = {V'_1}{f_i}\end{equation}\label{eq:map}

In this paper, the extracted spatial and temporal features are analyzed separately, and the mapping matrix $P$ is obtained after the PCA transformation. To analyze the effect of different principal components, we also compute the first 10-dimensional principal component transformation matrix $V_{10}'$, the size is $D \times 10$. For each dimension, we get the mapped vector separately, the visualization results are shown in Fig. \ref{fig:PCAf}. From this figure, we can see that the first-dimensional feature includes more information on the motion characteristics of repetitive actions. Therefore, in this paper, the first-dimensional principal component is used to count the repetitive action.

\subsection{Periodicity Mining and Repetition Counting}
Due to the complexity and diversity of the videos captured in the real scene and the non-standardization during the action execution, the principal component contains lots of noises. As shown in Fig. \ref{fig4a}, although there are some repetitive motion rules in the figure, the lower peak and the noises may lead to poor performance when counting. To locate the repetitive action, we need to distinguish the interesting actions from the unrelated noises. As discussed above, the interesting repetitive actions usually have two features. One is that it usually carries more energy than the other movement. The other is that it usually has a relatively higher frequency compared with the occasional camera motion. Therefore, we propose a high-energy-based repetitive action location method. And the counting is realized using the location results. In order to locate the high-energy repetitive action, we first give the frequency analysis of the repetitive action and then design a multi-stage filtering scheme.
\subsubsection{Frequency Analysis of the repetitive action}
We use the Fourier transform to analyze the repetition of the action. Fourier transform is usually used to estimate the power spectrum.  Specifically, the time-varying signal is decomposed into the superposition of the components in the frequency domain by Fourier transform. The vibration frequency of the signal is separated to get the spectrum by equation (5),
 where $P_u$ is the first-dimensional principal component of the $u$th frame, $N$ is the total number of frames for a video.\\

\begin{small}\begin{equation}{X(k) = \sum\nolimits_{u = 1}^N {P_u*exp\left( { - j*2\pi *(k - 1)*\frac{u}{N}} \right)} }\end{equation}\end{small}\label{eq:fft}

\begin{small}\begin{equation}X{(k)_{threshold \le k \le \left( {L - threshold} \right)}} = 0\end{equation}\end{small}\label{eq:threshold}

\begin{small}\begin{equation}P_u = \frac{1}{N}{\rm{ }}\sum\nolimits_{k = 1}^N {X(k)*exp\left( {j*2\pi *(k - 1)*\frac{{u - 1}}{N}} \right)} \end{equation}\end{small}\label{eq:ifft}

\begin{center}
\begin{figure}
\centering
\setcounter{subfigure}{0}
\subfigure[The first 1D of the component of the PCA]{\label{1PCA}\includegraphics[height=1in,width=0.45\textwidth]{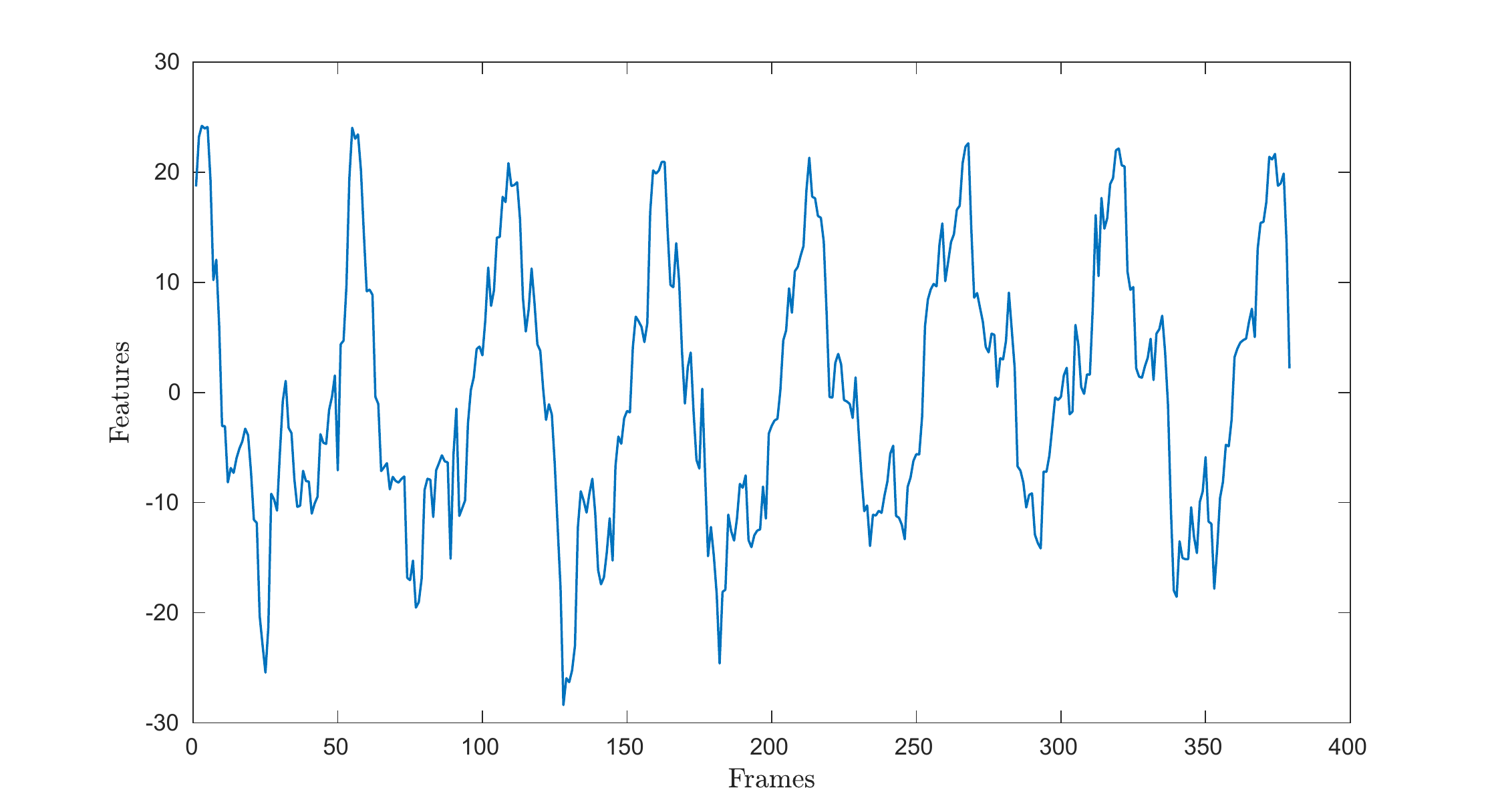}}
\subfigure[The spectrum obtained by the Fourier transform]{\label{FFTPCA}\includegraphics[height=1in,width=0.45\textwidth]{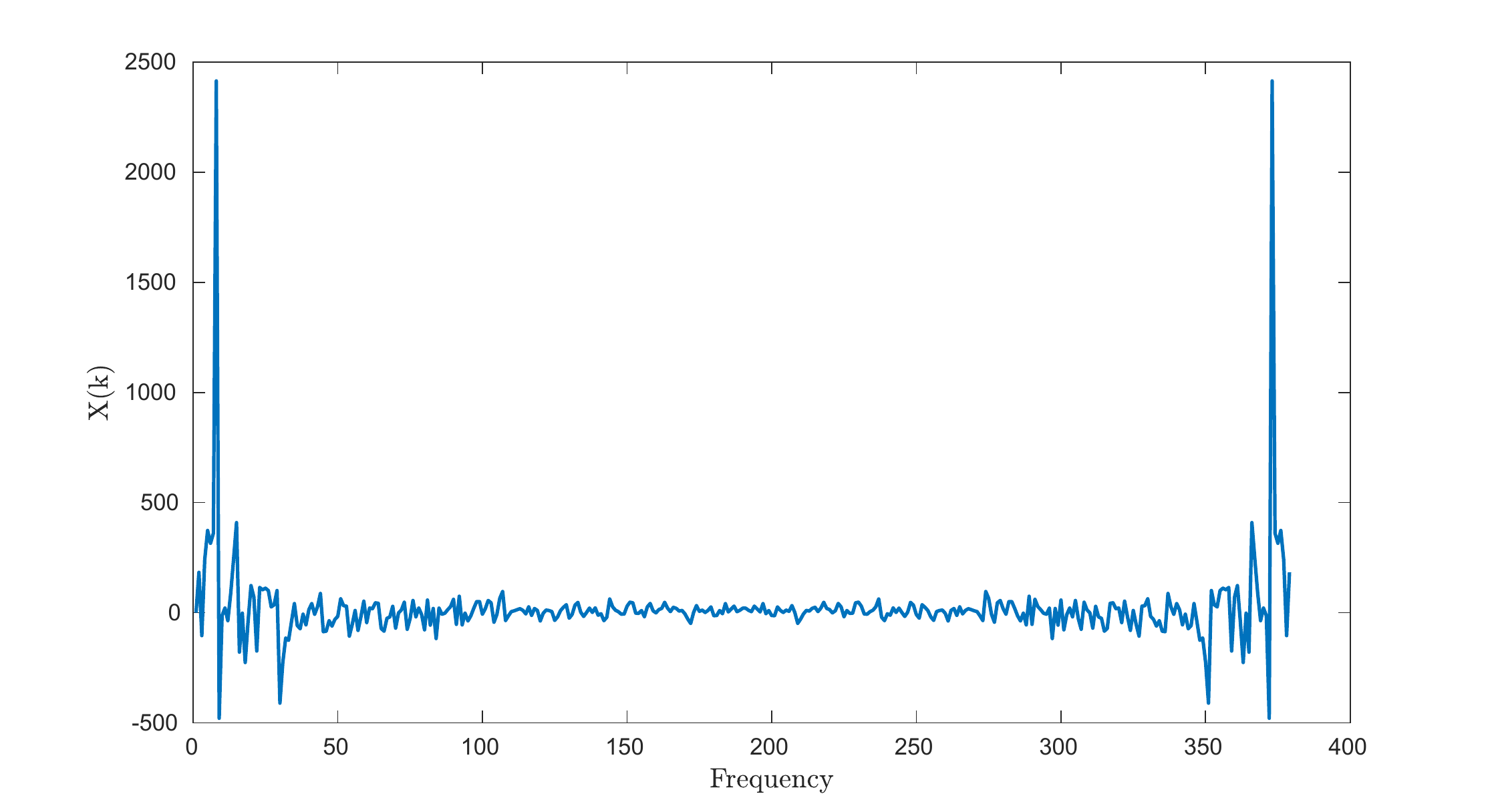}}
\subfigure[The principal component waveform obtained by Filtering and Inverse Fourier transform
]{\label{Filter}\includegraphics[height=1in,width=0.45\textwidth]{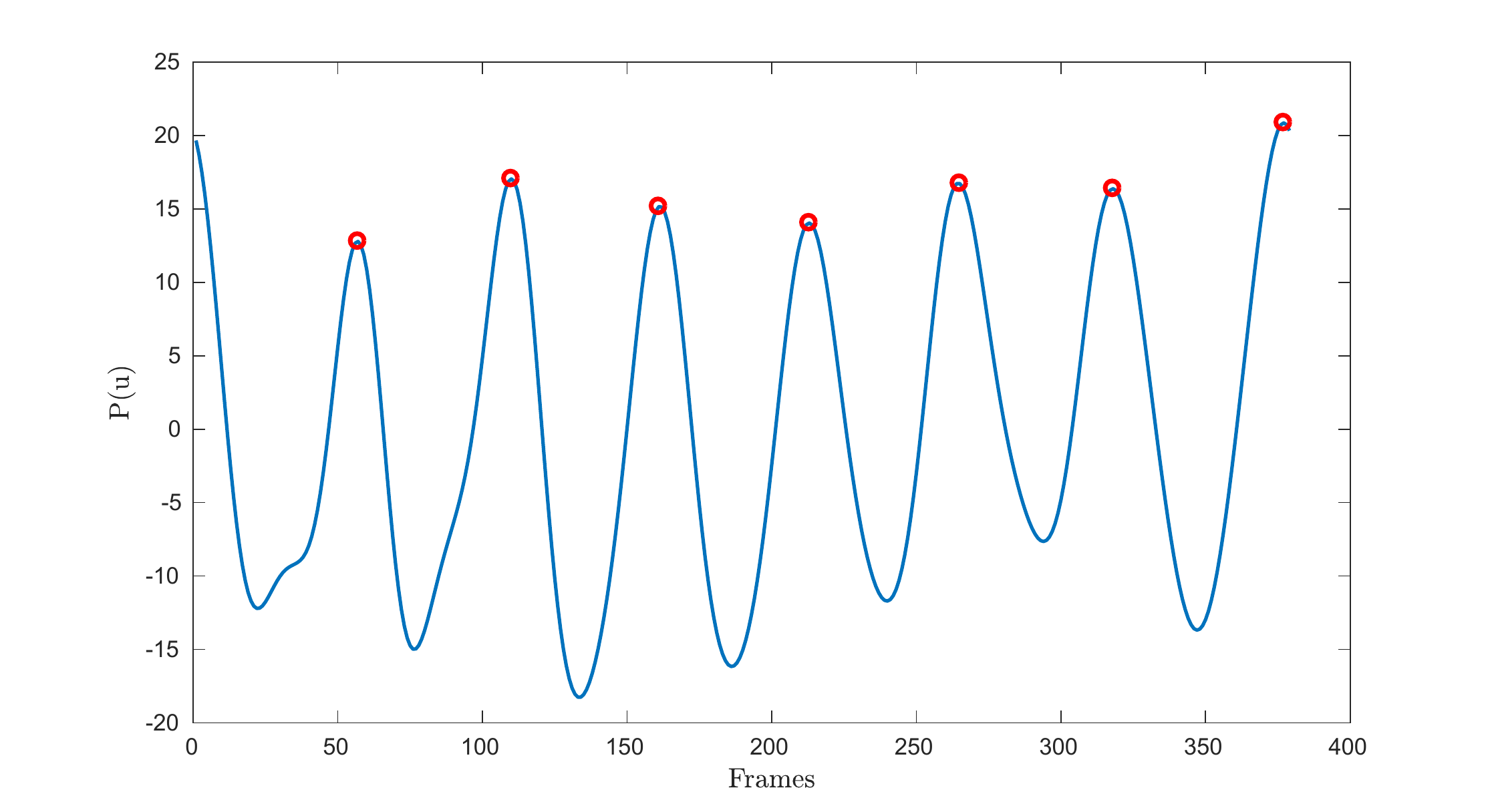}}

\caption{Repetitive action characteristics over time}
\label{fig:fPCA}    
\end{figure}
\end{center}

Fig. \ref{fig:fPCA} is the visualization of the principal component of a repetitive action video, where Fig. \ref{1PCA} is the first dimensional principal component reflecting the repetitive motion, and Fig. \ref{FFTPCA} gives its corresponding Fourier results. From Fig. \ref{1PCA}, we can see that there are lots of noises in the 1D waveform.  In other words, there are some other signals besides repetitive actions. Obviously, there are some unrelated actions or other motions in the video. To locate the repetitive action, we use the coefficients of the different frequency components to determine the filtered frequency band, and set the values of their corresponding frequency domain to 0, as in equation (6). 
 Then the original signal $P_u$ is obtained through the inverse Fourier transform by the equation (7). 
 The above method is used to filter the signal noise, so that the frequency graph has a smooth trajectory with repetitive motion rules, as shown in Fig. \ref{Filter}. Finally, the counting results can be obtained by simple peak detection.

\subsubsection{Fourier Transform with Multi-stage Threshold Filtering Analysis}
\begin{center}
\begin{figure}
\centering
\setcounter{subfigure}{0}
\subfigure{\label{Fig6apca}\includegraphics[height=1in,width=0.155\textwidth]{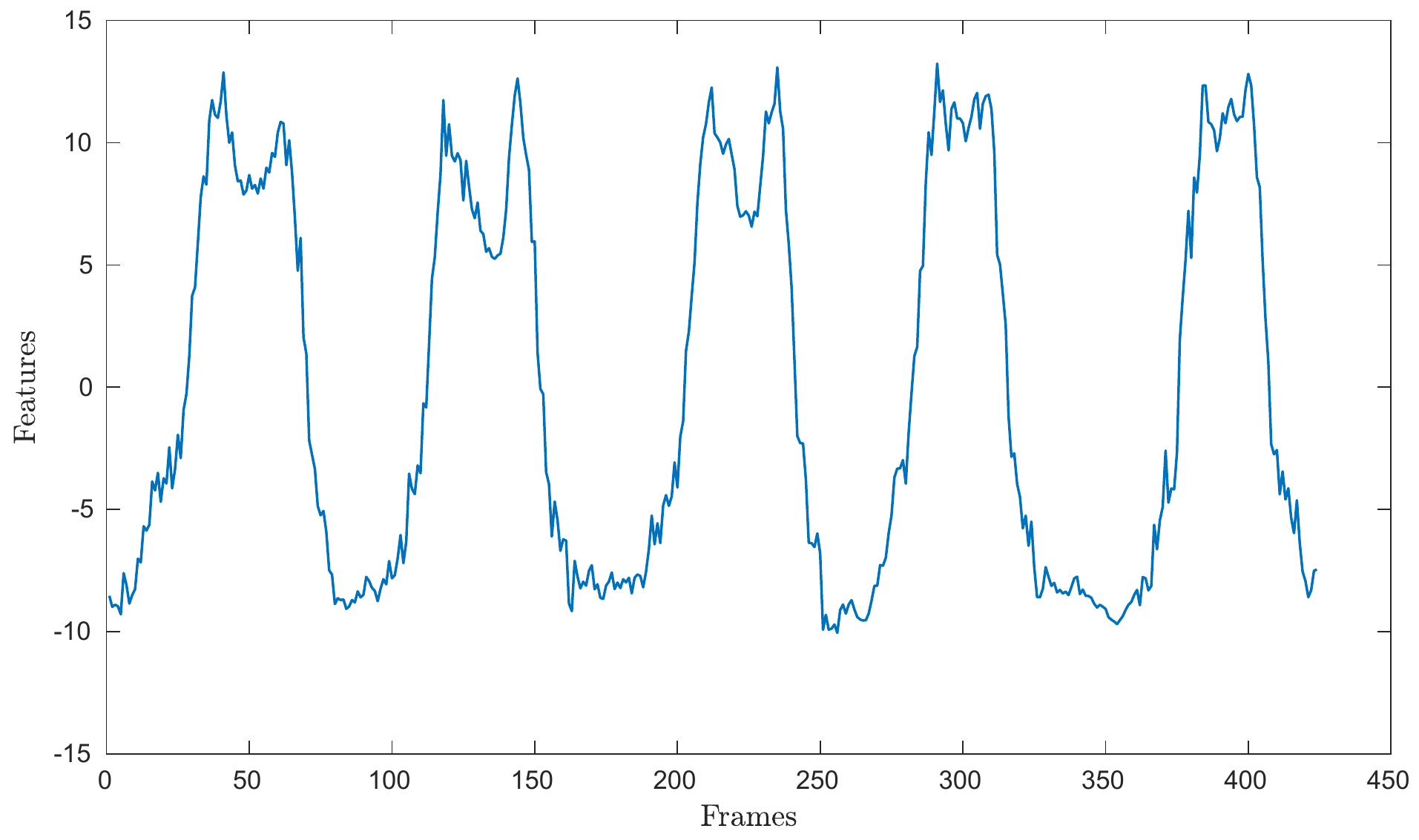}}
\subfigure{\label{Fig6bpca}\includegraphics[height=1in,width=0.155\textwidth]{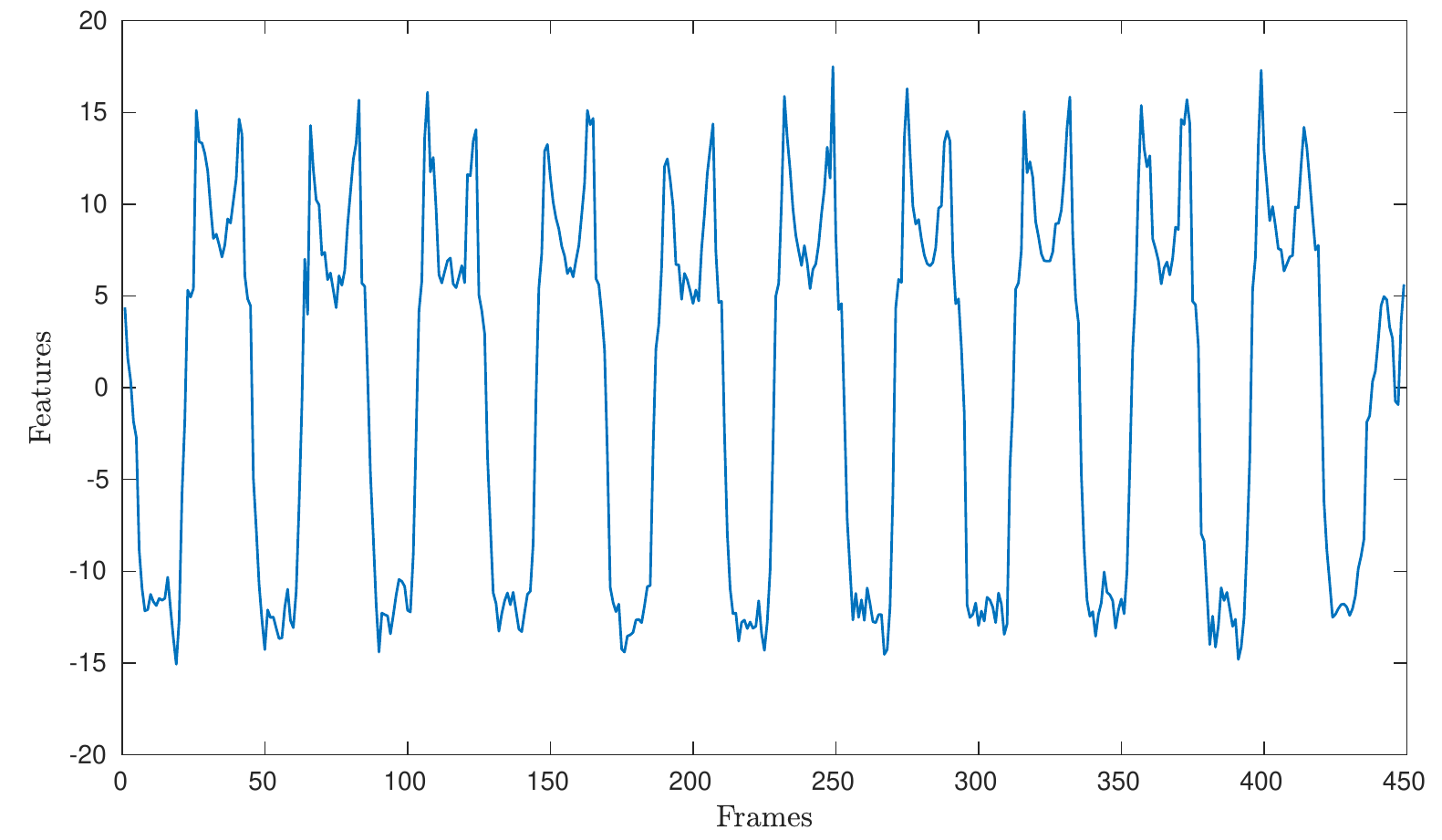}}
\subfigure{\label{Fig6Cpca}\includegraphics[height=1in,width=0.155\textwidth]{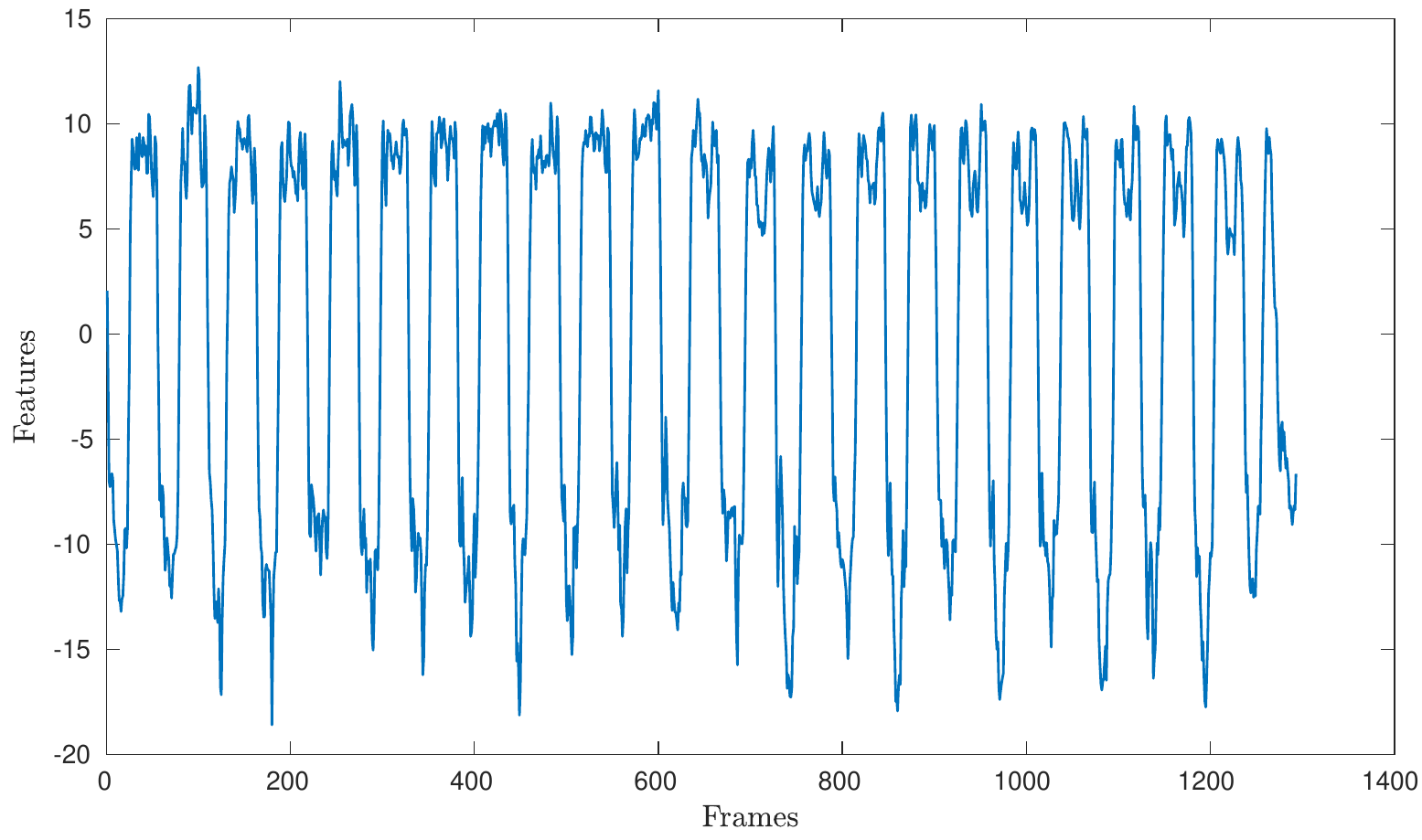}}
\subfigure{\label{Fig6a}\includegraphics[height=1in,width=0.155\textwidth]{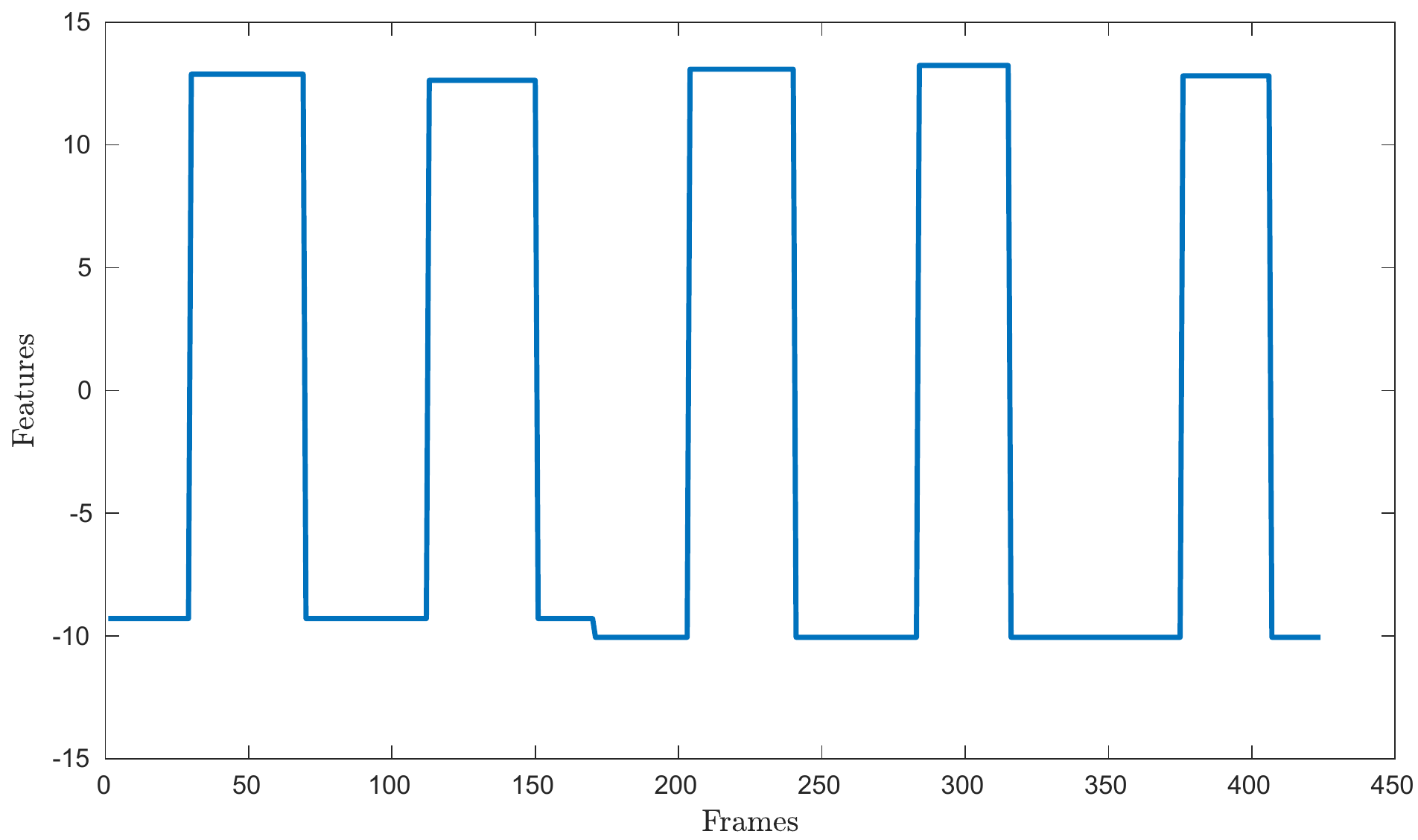}}
\subfigure{\label{Fig6b}\includegraphics[height=1in,width=0.155\textwidth]{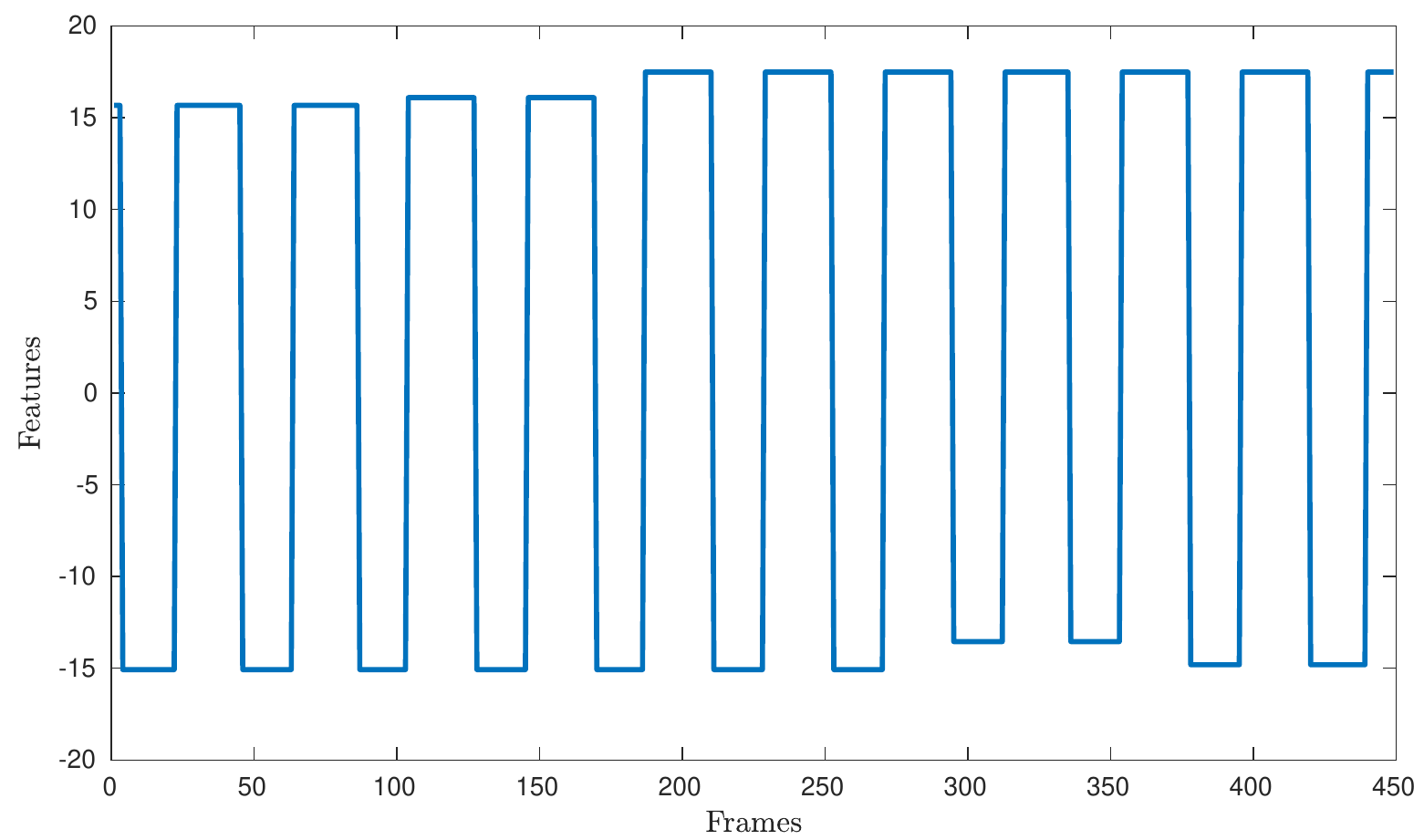}}
\subfigure{\label{Fig6c}\includegraphics[height=1in,width=0.155\textwidth]{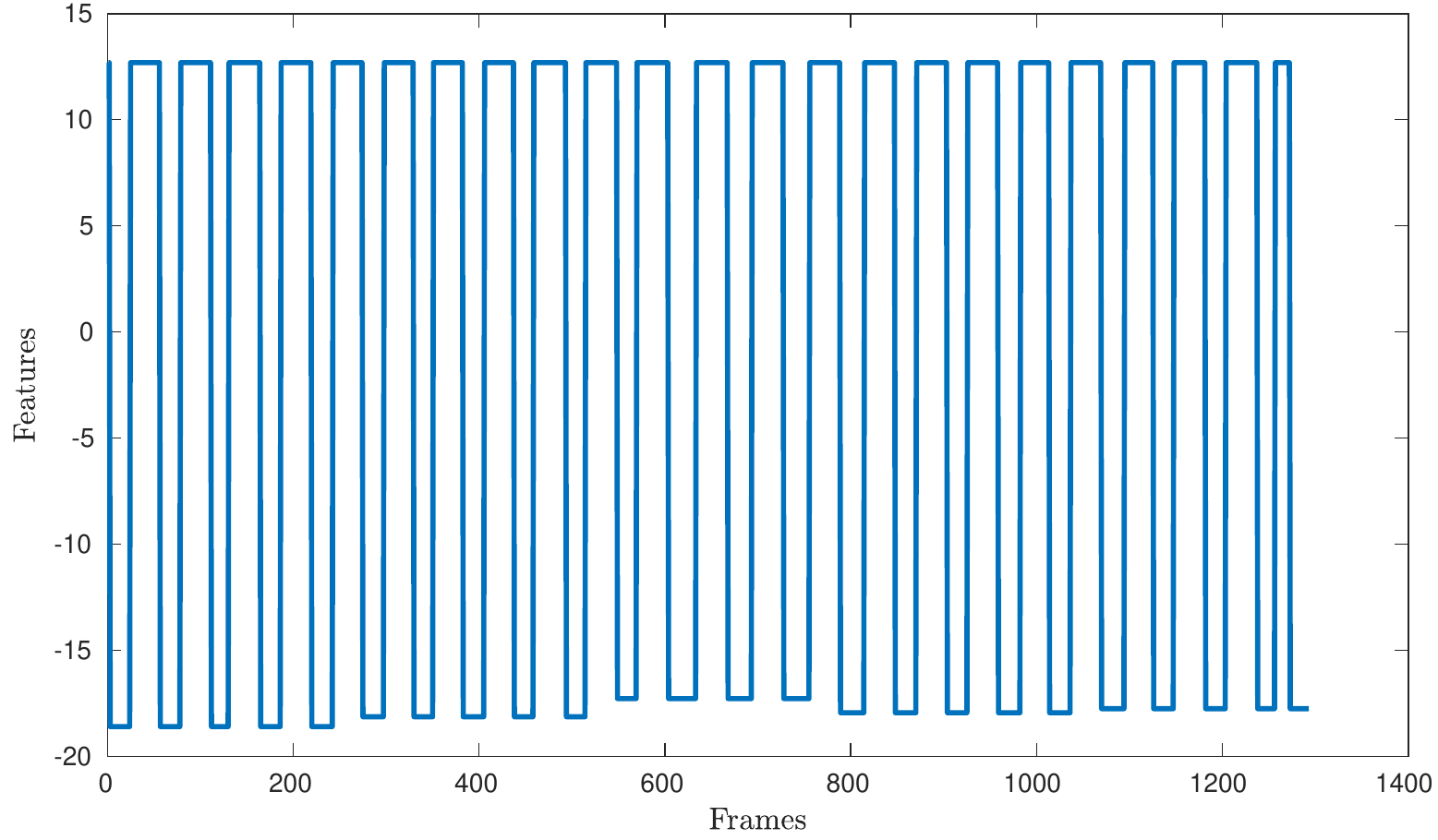}}
\caption{The number of the high-pass frequency band}
\label{fig:hpfb}    
\end{figure}
\end{center}

\begin{center}
\begin{figure}
\centering
\includegraphics[width=2in]{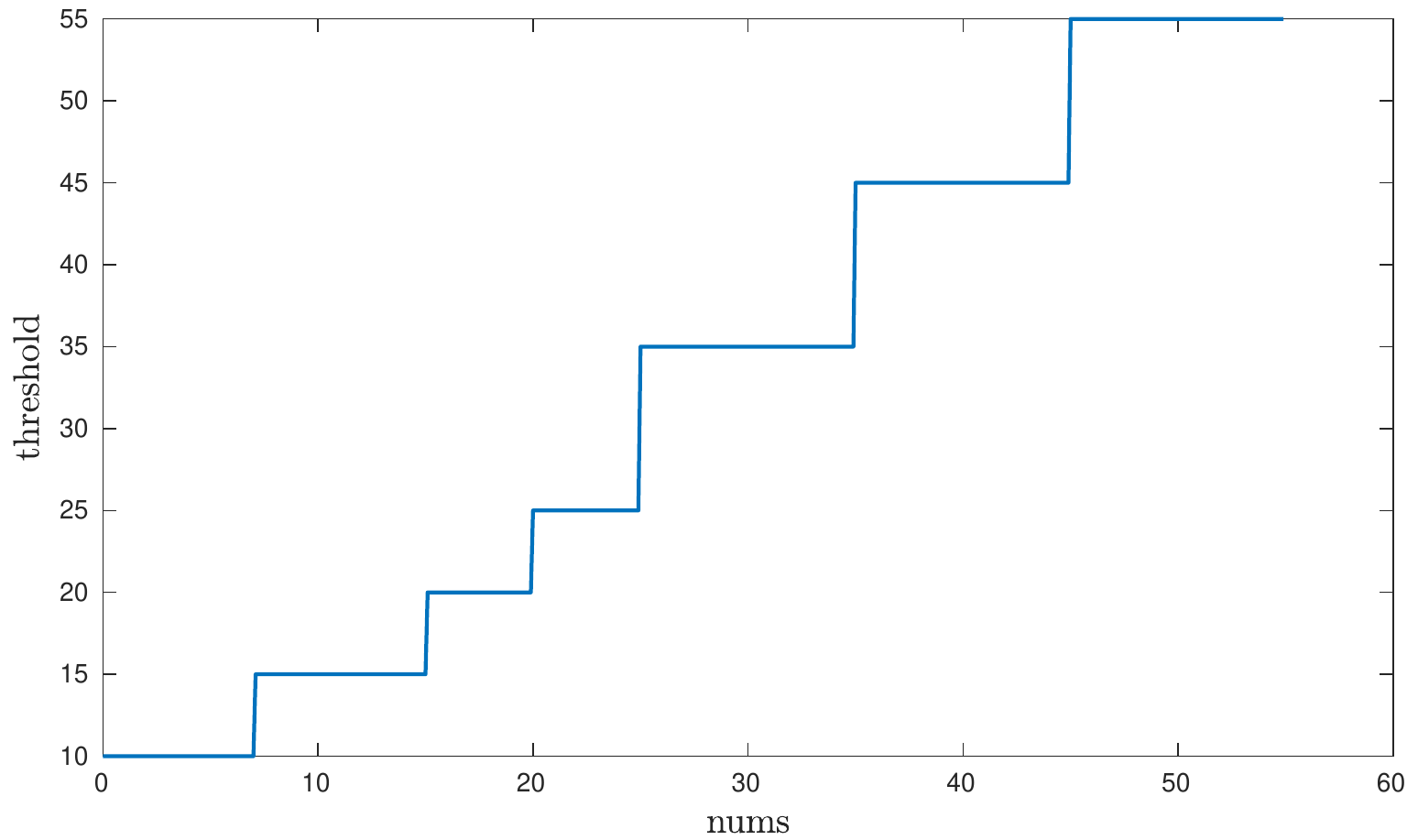}
\caption{The multistage thresholds}
\label{figmulti}
\end{figure}
\end{center}

In order to find the irrelevant frequency band discussed in the above section, we propose a filtering scheme, Fourier transform with Multi-stage
threshold Filtering(FMF). To obtain the periodic waveform well reflecting the times of the repetitive action, we need to filter the noises and unrelated actions. The key is to locate the interesting frequencies and delete the unrelated frequencies. The difference between the interested repetitive motion and the unrelated motion lies in its frequency and its corresponding energy. Obviously, the noise often has high frequency and low energy, while the interested repetitive action may often have a relatively low frequency and large energy. However, different repetitive actions may have different frequencies. Therefore, the filtering algorithm must adaptively adjust the filtering frequency threshold according to the motion characteristics of different frequencies, and we name our filtering algorithm as FMF. In order to do this, the number of the high-pass frequency band is calculated based on the first-dimensional principal component, as shown in Fig. \ref{fig:hpfb}. And according to the number, different thresholds are selected using different thresholds shown in Fig. \ref{figmulti}, which is empirically obtained by analyzing the effect of different motions on noises and uses different thresholds according to different frequencies of motions. Then the noise is removed by the spectrum obtained by the inverse Fourier transform, like \ref{Filter}. By the multi-stage benchmark, we modify the motion waveform and make the repetitive motion have obvious temporal periodic characteristics, which is useful to the counting of repetitive actions.

\section{Experiments}
YT\_Segments \cite{ref21} and QUVA \cite{ref22} datasets from Youtube are used to evaluate our algorithm. These data are diverse and challenging, and they also include the camera and background movement. The repetitive actions have varying lengths and complex appearance patterns.
\subsection{Datasets}
\textbf{YT\_Segments.}: It contains 100 videos with repetitive actions, including exercise, cooking, architecture, biology, and so on. To create a clean benchmark from very diverse videos, which are pre-split and only contain repetitive actions. The number of repetitive actions is pre-labeled per video. The smallest and largest numbers of the repetition are 4 and 50, respectively. The average duration of one video is 14.96s. Meanwhile, there are 30 videos with varying degrees of camera movement.\\
\textbf{QUVA.}: It's also made up of 100 videos and shows various kinds of repetitive video dynamics, including swimming, stirring, cutting, carding, and music-making. Compared with the YT\_Segments dataset, it has more challenges in cycle length, motion appearance, camera motion, and background complexity. Therefore, the dataset is a more realistic and challenging benchmark for estimating repetitive action.
\subsection{Evaluation Metrics and Baselines}
\textbf{Metrics.} We use the same evaluation criteria \cite{ref21} as those that used in the baselines as the metric for this task. For every video, the percentage of the absolute difference between ground truth $G$ and the predicted value $R$ is used: $\frac{{\left| {G - R} \right|}}{G} \times 100$. For $N$ videos, we calculate the Mean Absolute Error(MAE)
$ \pm $ standard deviation ($\sigma$) \cite{ref21} as the evaluation metrics, where 

\begin{equation}MAE = \frac{1}{N}\sum\limits_{i = 1}^N {\frac{{\left| {{G_i} - {R_i}} \right|}}{{{G_i}}}} \end{equation}
\begin{small}\[{\sigma ^2} = \frac{1}{N}\sum\nolimits_{i = 1}^N {{{\left( {G - R} \right)}^2}} \]\end{small}

\textbf{Baselines.} We compare our method with one classical method \cite{ref5} and two recent methods in \cite{ref21} \cite{ref22}. When reporting the results, we directly make use of the results provided in the paper \cite{ref22}.
\subsection{Results}

\textbf{Different thresholds and FMF.} Using spatial features, we test the repetition counting results with different thresholds and FMF. At first, we test the results of different thresholds, as shown in Table 1. We set the different thresholds empirically as $\alpha={10,15,20,25,30,35,\cdots}$. On the YT\_Segments dataset, the action repetition counting experimental results are shown in Table 1. From Table \ref{thres}, we can see that the results of the single threshold are much worse than the multi-stage thresholds. Because the repetitive actions have different frequencies, the interested frequency in one video may become noise frequency in the other video.
And we can also find that the relatively higher accuracy can be obtained by threshold 15. We think the possible reason is that lots of actions may have this frequency. However, for some high-frequency actions, this threshold will filter the interested frequency out, which makes that the multi-stage threshold performs better.\\

\begin{table}[!t]
\renewcommand{\arraystretch}{1.3}
\caption{Comparative Results of different thresholds }
\label{thres}
\centering
\begin{tabular}{c||c}
\hline\toprule[1pt]
\bfseries $\alpha $&  MAE  \\
\hline\hline
10 & 23.3 $\pm$ 63.4\\\hline
15&	19.9 $\pm$ 42.8\\\hline
20&	30.9 $\pm$ 38.14\\\hline
25	&44.6 $\pm$ 42.8\\\hline
30	&56.0 $\pm$ 62.7\\\hline
35	&67.2 $\pm$ 91.6\\\hline
multi-stage $\alpha$ &\bfseries 8.7 $\pm$ \bfseries 3.9\\\hline\toprule[1pt]

\end{tabular}
\end{table}

\begin{center}
\begin{figure}
\centering
\setcounter{subfigure}{0}
\subfigure[The counting results on YT\_Segments]{\label{YTR}\includegraphics[height=1in,width=0.5\textwidth]{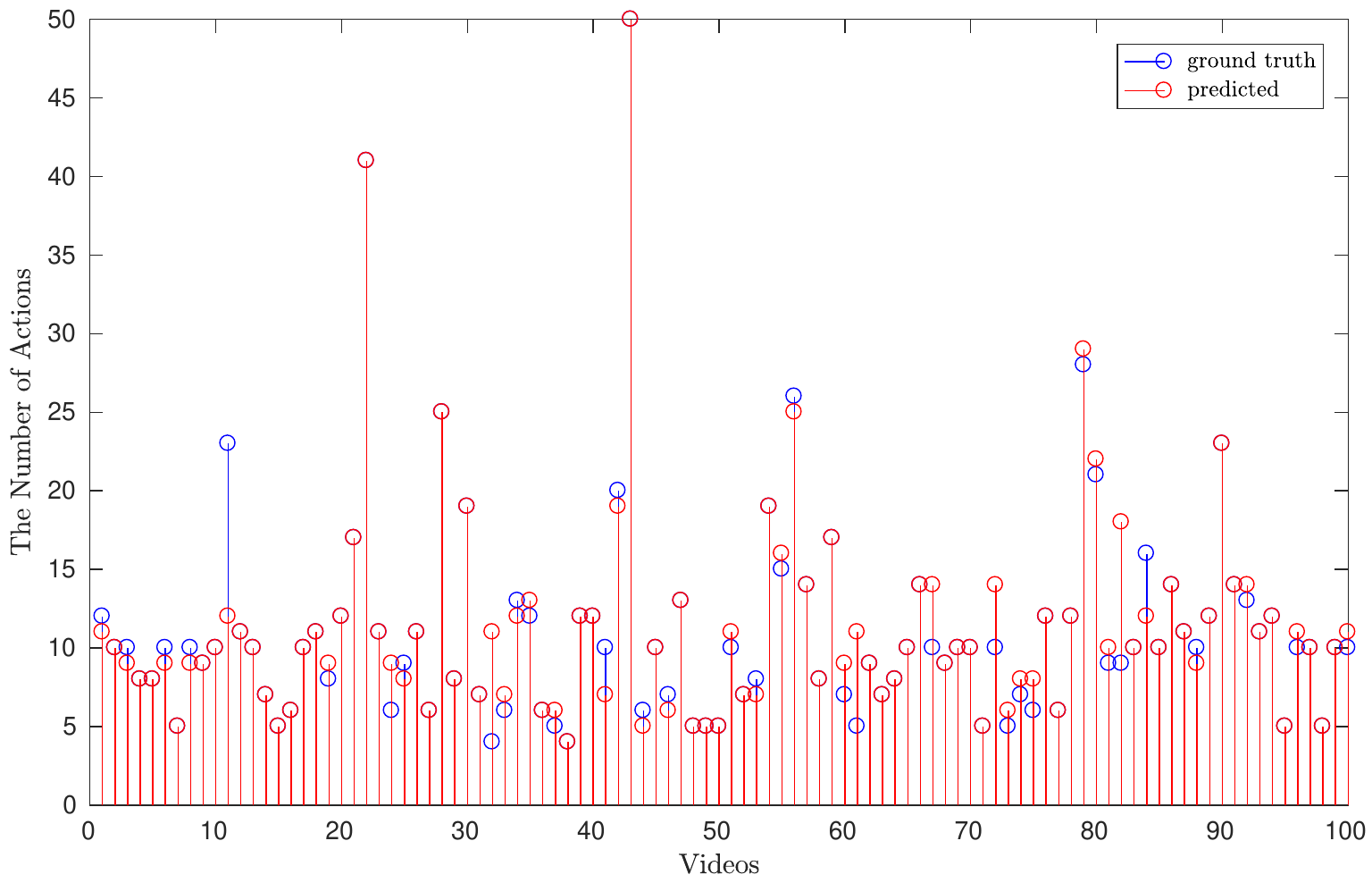}}
\subfigure[The counting results on QUVA]{\label{QUVAR}\includegraphics[height=1in,width=0.5\textwidth]{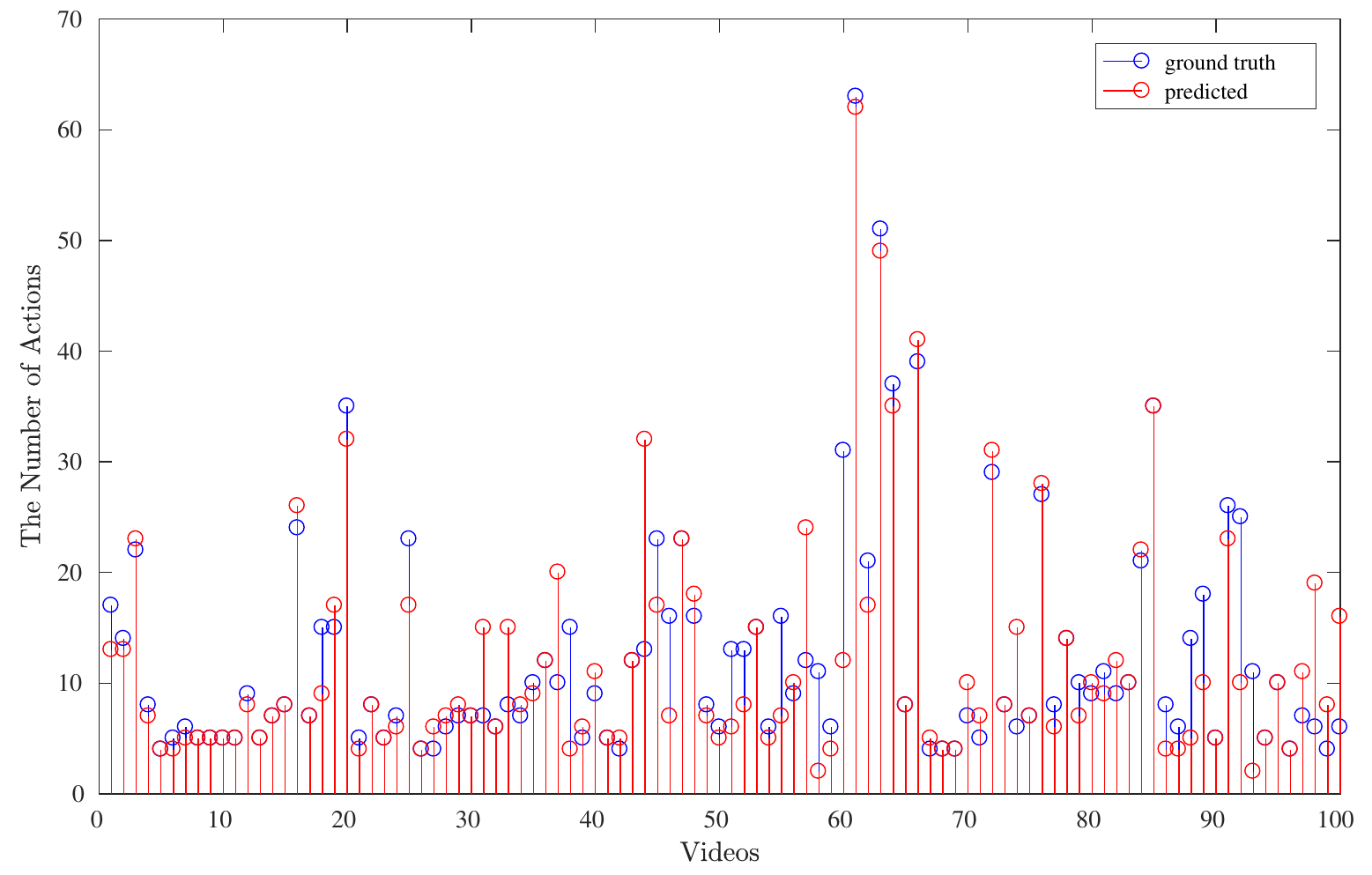}}
\caption{The detailed counting results of our method}
\label{fig:detailR}    
\end{figure}
\end{center}

\begin{center}
\begin{figure}
\centering
\includegraphics[width=3.3in]{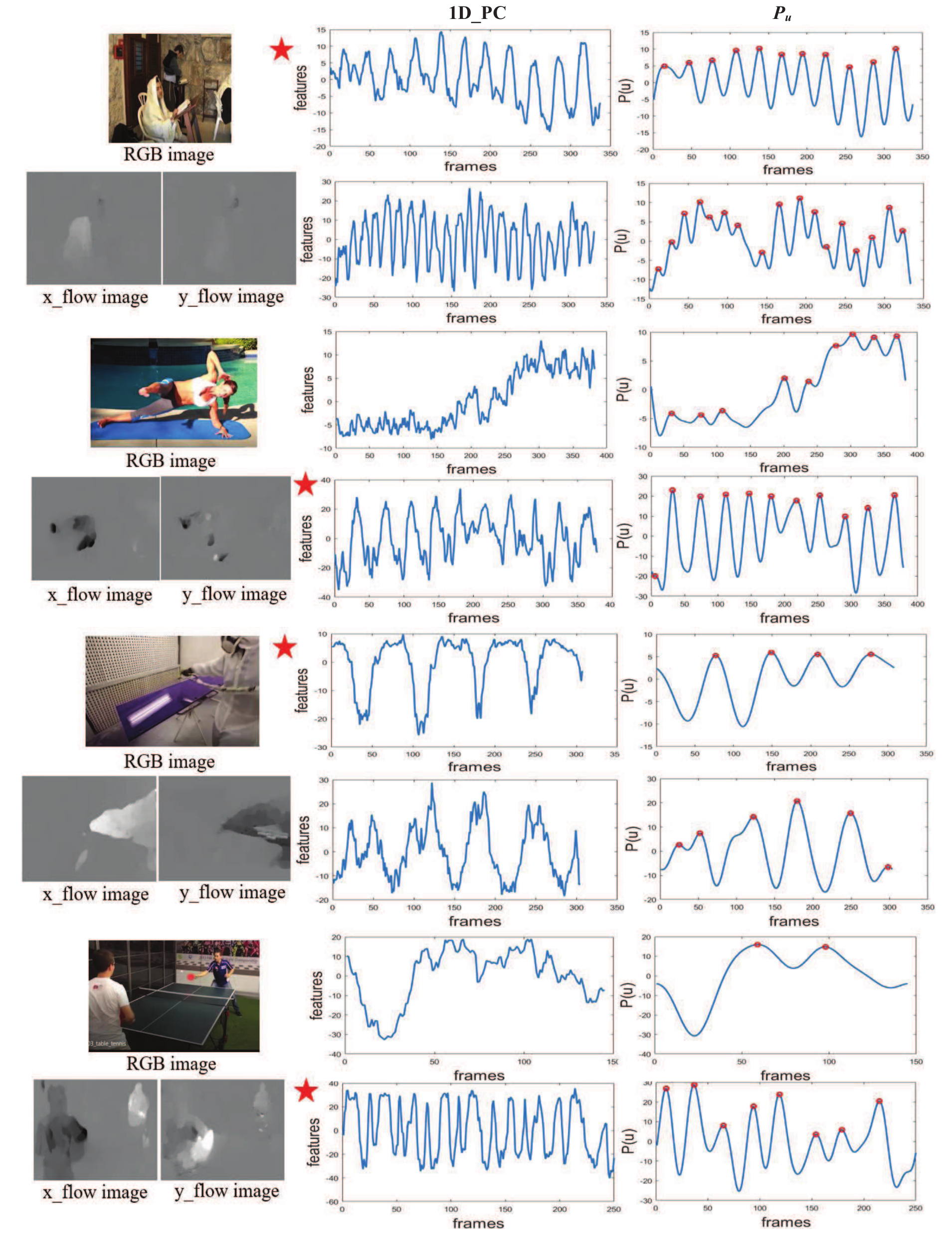}
\caption{The comparison of counting results based on RGB features and optical flow features, where 1D PC is the first dimensional principal component by PCA and the corresponding result by FMF based on the 1D PC over the time sequence. The position of Red star indicates the optimal detection result.}
\label{figqu}
\end{figure}
\end{center}

  Using spatial features, temporal features and the fusion features of spatial and temporal features, we further compare the repetition counting performance using FMF module and without FMF, as shown in Table \ref{expC}. And the influence of FMF module of our method is also analyzed.. From Table \ref{expC}, we can see that the accuracy is improved significantly by adding FMF module on the YT\_Segments dataset. It also shows that the spatial features based on RGB achieve the highest accuracy on the YT\_Segments dataset.
\begin{table}[!t]
\renewcommand{\arraystretch}{1}
\caption{Experimental Results Comparison about Whether Remove FMF on YT\_Segments Dataset}
\label{expC}
\centering
\setlength{\tabcolsep}{1mm}{
\begin{tabular}{c|c|c|c|c|c}\hline\toprule[1pt]
\multicolumn{3}{c|}{MAE Without FMF} & \multicolumn{3}{|c}{MAE With FMF} \\\hline

RGB & FLOW & RGB  FLOW  & RGB & FLOW & RGB  FLOW\\\hline
 13.7$\pm$  6.2 &  29.2$\pm$  21.3&  18.2 $\pm$  9.4 &\bfseries 8.7$\pm$\bfseries 3.9&21.9$\pm$ 12.7& 15.8$\pm$\ 6.6\\\hline\toprule[1pt]
\end{tabular}}
\end{table}
\textbf{Comparisons with state-of-the-art baselines.} We compare the performance of our method with state-of-the-art baselines. For YT\_Segments and QUVA, the spatial features and temporal features are used separately. At the same time, for convenience of comparison, we give the average of the methods on YT\_Segments and QUVA, and we name it Overall. The quantitative comparisons are presented in Table \ref{CompBaseline}. Compared with the existing baselines, our method achieves comparable performance without extra preprocessing. For the YT\_Segments dataset, the method of \cite{ref21} performs best with the MAE of 6.5. The MAE in \cite{ref22} is 10.3, which is better than the article \cite{ref5}. Our method is superior to articles \cite{ref5} \cite{ref22} with the MAE of 8.7, but the standard error achieves the best performance compared with the above methods, which illustrates that the worst performance of our method is the best in all the methods. Moreover, the results also show that our method can achieve good performance under the static background. In the more challenging QUVA dataset, our experimental results also achieved decent performance. The method \cite{ref21} performed the worst with the MAE of 48.2, because their network considered only four types of action during training. The method of \cite{ref5} was 38.5. In \cite{ref22}, the MAE was 23.2. The performance of our method also comes in the second place, which is very close to the best performance. And the standard error is also the least. These results show that our method can also adapt well to the dynamic background. In summary, compared to the above methods, we get the best standard error on the two public datasets. At the same time, we get the MAE very close to the state-of-the-art baseline. The results show that our method can achieve comparable results in counting action repetition for unconstrained videos with a decent framework, not relying on preprocessing.\\
%
\begin{table}
\caption{Comparisons with the state-of-the-art baselines}
\label{CompBaseline}
\begin{tabular}{l|c|c|c}\hline\toprule[1pt]

\centering
 \diagbox{Methods}{Datasets}& YT\_Segments & QUVA &Overall\\\hline
 Pogalin et al. \cite{ref5}&21.9$\pm$  30.1 &  38.5$\pm$  37.6 &30.2$\pm$33.9\\\hline
 Levy \& Wolf \cite{ref21}& {\bfseries {6.5}} $\pm$9.2&	48.2$\pm$61.5&27.4$\pm$35.4\\\hline
 Runia \&Snoek \cite{ref22}&10.3$\pm$19.8&	{\bfseries {23.2}}$\pm$34.4&\bfseries 16.8$\pm$27.1\\\hline
 Ours&8.7$\pm$\bfseries 3.9	&25.1$\pm$\bfseries 26.3&16.9$\pm$ \bfseries 10.4\\\hline\toprule[1pt]
\end{tabular}
\end{table}
\textbf{Detailed results and analysis.} To further validate our contribution, on YT\_Segments and QUVA, we give the counting results in detail, as shown in Fig. \ref{fig:detailR}. \\
 From Fig. \ref{YTR}, we can see that the counting results of most of the actions are very close to their groundtruth and the differences cluster around positive or negative 1. Our peak-detection counting scheme can well explain this. Because we use the number of the peaks as the counting results, it is slightly different from the repetitive case, where the repetition is, in fact, a cycle. Therefore, using more detailed cycle detection may solve this problem. In addition, there is only 1 sample (video 11) whose error is relatively large, and the corresponding video is a repetitive action with the sub-movements of the left-arm and right-arm. Our method regards one repetitive action, which includes the repetitive motion of the left-arm and right-arm as once. However, the groundtruth regards that the motion of the left-arm and right-arm arm are twice. We think our result is interpretable because these two movements are, in fact, two sub-movements of one movement. From this point, although our performance is not good enough, but the actual performance of our method is largely better than we gave.\\
 From Fig. \ref{QUVAR}, besides the above mentioned negative and positive 1 difference problem, there are also other problems. Some groundtruth results are almost twice as our results, for example, video 34. We found this error occurs because those actions often include two coupled actions; for example, in video 34, the whole action includes the left leg and right leg sub-movements. This will contribute 2 times of 1 action in our algorithm. The other error is due to the changing of the view angles of the action, like video 31. For these two problems, we think they can be solved using more advanced signal processing methods. And we will focus on it in our future work.\\
 In summary, the main error of our method lies in two aspects. One is the difference between the peak and the cycle. The other is the coupled submovements in action. Both kinds of errors become from the processing of the deep features. Therefore, this also validates our insight that the 1D principle component of the deep features can well model the periodicity of the repetitive action.
\textbf{Feature analysis.}
Fig. \ref{figqu} shows the visualization results of different features by our method on YT\_Segments and QUVA datasets. Only three parts are shown here, and the first part is the input data to get the deep features by BN-Inception Network, the second part is the sequence waveform of action obtained by PCA when $k=1$, and the third part is the final result for counting. The experimental results indicate that static features achieve better results in a relatively clean background, and results are poorer with serious background interference. However, the method based on optical flow features achieved prominent results. They demonstrate the effectiveness of the proposed method in this paper. Moreover, we also can find that the 1D principal component of the deep features includes good periodic information of the action. If we can further improve the signal processing after the periodicity mining, the performance can improve further.

%

\section{Conclusions}
We propose an important insight that the periodicity of the action can be well modeled by the deep features extracted from the action recognition task. We think this insight is very important for repetition counting due to two reasons. On the one hand, the repetition counting method can borrow the state-of-the-art results or experiences from action recognition, which decreases the gap between the development of action recognition and the repetition counting. On the other hand, this insight can simplify the repetition counting task from the trivial preprocessing or synthetic mode generation. \\
Based on this insight, we propose a new counting method using high energy rules for unconstrained video. Due to the introduction of energy, our method can solve unconstrained action modes in unconstrained videos. In detail, by using the training model based on Kinetics, we extract reliable deep features, including the temporal evolution characteristics of video actions and the unique appearance and spatiotemporal characteristics of motion patterns by deep ConvNets. Then, the periodic movement information can be obtained by the PCA based on the deep features of the action. Besides, we compute the frequency spectrum by Fourier transform to remove noise information by multi-stage threshold filtering. Then, the time sequence waveform is smoothed, and the action repetition counting task is completed according to peak detection. Extensive experimental results show the effectiveness of our method.\\
Compared with the existing methods, our method is simple and flexible without preprocessing. However, it still has poor performance when there is interference or chaotic background in the motion, especially when there are coupled repetitive actions and continuous changing of the viewpoints. These non-periodic interference motions make it impossible to analyze the motion characteristics of the target object accurately. We will focus on these problems in the future.

\ifCLASSOPTIONcompsoc
  \section*{Acknowledgments}
\else
  \section*{Acknowledgment}
\fi

This work was supported in part by the National Natural Science Foundation of China under Grant 61673192, U1613212,  and in part by the Basic Scientific Research Project of the Beijing University of Posts and Telecommunications under Grant 2018RC31.

\ifCLASSOPTIONcaptionsoff
  \newpage
\fi



%

%

\bibliographystyle{IEEEtran}
\end{document}